\documentclass[numbers,trsc, nonblindrev]{informs3} % current default for manuscript submission
\usepackage{lmodern}
\usepackage{fix-cm}
\usepackage[numbers]{natbib}
\usepackage{graphicx}
\usepackage{textcomp}
\usepackage{subfigure}
\usepackage{url}
\usepackage{verbatim}
\usepackage{graphicx}
\usepackage{amssymb}

\newcommand{\qedsymbol}{\ensuremath{\square}} 
\newenvironment{myproof}[1][Proof]{\textit{#1.}\ }{\hfill \qedsymbol}

\usepackage{array,booktabs}
\usepackage{multirow}
\usepackage{bm}
\usepackage{bbding}
\usepackage{pifont}
\usepackage{makecell}
\usepackage[backref]{hyperref} 
\hypersetup{hidelinks}
\usepackage{multirow}
\usepackage{xcolor}
\usepackage{array}
\usepackage{tabularx}
\usepackage{newfloat}
\usepackage{listings}
\usepackage{multicol}
\usepackage[linesnumbered, ruled]{algorithm2e}

\OneAndAHalfSpacedXI % current default line spacing
\usepackage{natbib}
 \bibpunct[, ]{(}{)}{,}{a}{}{,}%
 \def\bibfont{\small}%
\TheoremsNumberedThrough     % Preferred (Theorem 1, Lemma 1, Theorem 2)
\EquationsNumberedThrough    % Default: (1), (2), ...

\begin{document}
%%%%%%%%%%%%%%%%

% Title or shortened title suitable for running heads. Sample:
\RUNTITLE{Liao et al., Human-Like Trajectory Prediction for AD}
% Enter the (shortened) title:
%\RUNTITLE{}

% Full title. Sample:
\TITLE{Towards Human-Like Trajectory Prediction for Autonomous Driving: A Behavior-Centric Approach}

\ARTICLEAUTHORS{
\AUTHOR{Haicheng Liao}
\AFF{State Key Laboratory of Internet of Things for Smart City and Department of Computer and Information Science, University of Macau}%, \EMAIL{}, \URL{}
\AUTHOR{Zhenning Li\thanks{Corresponding author.} }
\AFF{State Key Laboratory of Internet of Things for Smart City and Departments of Civil and Environmental Engineering and Computer and Information Science, University of Macau, \EMAIL{zhenningli@um.edu.mo}}

\AUTHOR{Guohui Zhang}
\AFF{Department of Civil and Environmental Engineering, University of Hawaii at Manoa}

\AUTHOR{Keqiang Li}
\AFF{School of Vehicle and Mobility, Tsinghua University, Beijing, China}

\AUTHOR{Chengzhong Xu}
\AFF{State Key Laboratory of Internet of Things for Smart City and Department of Computer and Information Science, University of Macau}%, \URL{}
} % end of the block

\ABSTRACT{Predicting the trajectories of vehicles is crucial for the development of autonomous driving (AD) systems, particularly in complex and dynamic traffic environments. In this study, we introduce HiT (Human-like Trajectory Prediction), a novel model designed to enhance trajectory prediction by incorporating behavior-aware modules and dynamic centrality measures. Unlike traditional methods that primarily rely on static graph structures, HiT leverages a dynamic framework that accounts for both direct and indirect interactions among traffic participants. This allows the model to capture the subtle, yet significant, influences of surrounding vehicles, enabling more accurate and human-like predictions. To evaluate HiT’s performance, we conducted extensive experiments using diverse and challenging real-world datasets, including NGSIM, HighD, RounD, ApolloScape, and MoCAD++. The results demonstrate that HiT consistently outperforms state-of-the-art models across multiple metrics, particularly excelling in scenarios involving aggressive driving behaviors. This research presents a significant step forward in trajectory prediction, offering a more reliable and interpretable approach for enhancing the safety and efficiency of fully autonomous driving systems.
}%

% Fill in data. If unknown, outcomment the field
\KEYWORDS{Autonomous Driving, Trajectory Prediction, Driving Behavior, Interaction Understanding}
% \HISTORY{}

\maketitle

\section{Introduction}

The success of autonomous vehicles (AVs) hinges on their ability to navigate complex and unpredictable driving environments with the same level of insight and adaptability as human drivers \citep{li2024context,liao2024human}. Central to this challenge is the need for AVs to predict the trajectories of surrounding vehicles with a level of sophistication that mirrors human intuition. Unlike machines, human drivers anticipate the actions of others by drawing on subtle behavioral cues—such as slight changes in speed or lane positioning—and making split-second decisions based on a deep understanding of context and intent \citep{schwarting2019social}. This human-like capability to interpret and respond to the dynamic and often ambiguous behaviors of other drivers is essential for safe and efficient driving. Replicating this ability in AVs remains one of the most formidable tasks in autonomous driving (AD) research \citep{feng2023dense,liao2024gpt}.

As we strive to elevate the trajectory prediction capabilities of AVs, the incorporation of human-like abilities to comprehend and appropriately respond to their surroundings may prove to be a viable solution. As highlighted in \citep{schwarting2019social}, accounting for the behaviors of other drivers in the decision-making processes of AVs can potentially result in enhanced driving performance. Building on this premise, we advocate that a deeper dive into driver behavior can significantly uplift trajectory prediction for AVs. Wang et al. \citep{wang2022social} have proposed several key inquiries pertaining to how AVs can comprehend the interactions and behaviors of other vehicles, including but not limited to, when interactions occur, who is involved in these interactions, and how can we quantitatively evaluate and analyze the social dynamics at play in these interactions? The tireless efforts of our predecessors in the field have resulted in the illumination of the responses to the initial two inquiries \citep{deo2018convolutional}, yet the final query remains a conundrum, particularly in regards to identifying the behaviors of other drivers and determining the ramifications of such behaviors on the execution of driving maneuvers.

Despite advancements in technology, current trajectory prediction models frequently fall short when confronted with the complexities of real-world traffic.  Most existing research \citep{schwarting2019social,chandra2020cmetric} on modeling driving behavior has been conducted in a large amount of manual label annotations in the training process, which can be time-consuming and costly to collect. Moreover, these models often rely on predefined behavior categories and simplistic assumptions about interactions between vehicles, which do not adequately capture the continuous, context-sensitive nature of driving behavior \citep{zhao2024human}. For instance, a driver’s decision to change lanes is influenced by a myriad of factors, including the behavior of nearby vehicles, road conditions, and overall traffic flow—factors that are not easily reduced to rigid categories. Furthermore, selecting an appropriate time window for behavior recognition poses additional challenges. These inflexible methods are prone to fluctuations that significantly reduce detection accuracy, making them unsuitable for long-term trajectory prediction. This lack of behavioral nuance in traditional models often leads to inaccurate predictions and suboptimal decision-making by AVs.

Another critical challenge lies in addressing the inherent uncertainty and variability in human driving behavior. Human drivers make decisions based on a complex interplay of factors that are often difficult to predict with certainty \citep{li2024steering}. Traditional models, which rely on rigid, predefined states, struggle to accommodate the full spectrum of possible behaviors, particularly in real-time settings where computational resources are limited. This challenge underscores the need for a more flexible and adaptive approach to trajectory prediction that can better reflect the unpredictability and complexity of human behavior \citep{liao2025cotdrive}.

Previous investigations have posited that there exists a certain relationship between different drivers' behaviors and their driving performance \citep{yu2024online}. When confronted with the prospect of another vehicle attempting to overtake, drivers who exhibit more aggressive inclinations may opt to invigorate their speed, with the intent to impede the passing vehicle. On the other hand, conservative drivers may elect to slightly reduce their velocity, thereby affording greater latitude for the overtaking vehicle to pass with ease. Moreover, existing research \citep{wang2025nest} suggests that the driving behaviors of motorists on the road exhibit remarkable consistency, predictability, and persistence \citep{liao2024bat,wang2025dynamics}. For instance, a driver who habitually tailgates other vehicles is likely to persist in this behavior in the future, barring any significant changes in the driver's circumstances or environment. Conversely, a driver who consistently maintains a safe following distance is unlikely to suddenly exhibit reckless behavior and tailgate other drivers. By leveraging the stability and recurrence characteristics of driving behaviors, sophisticated trajectory prediction algorithms can be developed to enable AVs to anticipate the future movements of other vehicles with increased precision and accuracy. Therefore, crucial to prediction models is the comprehensive understanding of dynamic interactions between different road users, enabling them to accurately interpret and respond to human behaviors and intentions in mixed traffic scenarios.

In addition, as AVs operate in real-time environments, rapid and accurate decision-making is of paramount importance, underscoring the need for efficient trajectory prediction models. The trajectory prediction models need to be designed to provide fast and precise predictions, enabling timely responses to dynamic road conditions and ensuring vehicular and environmental safety \citep{wang2025wake,liao2024cdstraj}. However, autonomous vehicles often run on embedded devices with limited computational resources, necessitating models that maximize prediction accuracy whereas minimizing computational demands. This balance allows the autonomous driving system to concurrently perform other crucial tasks, including object detection, traffic sign recognition, and path planning.

In response to these challenges, we introduce \textbf{HiT} (Human-Like Trajectory Prediction), a novel model that advances the state of the art in trajectory prediction for AVs by addressing the critical limitations of existing approaches. HiT integrates a behavior-aware framework that transcends traditional rigid classifications by capturing the fluid and context-sensitive nature of human driving behavior, dynamically adapting to changes in driver actions based on centrality measures and principles from traffic psychology. To effectively model the complex and multi-agent interactions typical in dense traffic scenarios, HiT employs a lightweight hypergraph-based interaction framework, which goes beyond the limitations of traditional graph-based models by capturing higher-order interactions among multiple agents. Additionally, recognizing the challenges of unstructured environments, HiT adopts a polar coordinate system for spatial representation, enhancing its adaptability to irregular road topologies and ensuring accurate predictions across diverse driving conditions. The model further integrates the q-rung orthopair fuzzy weighted Einstein Bonferroni mean (q-ROFWEBM) operator to manage the inherent uncertainty and variability in human driving behavior, allowing for the processing of a broader range of behavioral information and improving prediction accuracy and robustness. Moreover, the adaptive nature of HiT reduces its dependency on extensive manual annotations, enabling the model to generalize more effectively across different traffic scenarios whereas maintaining high prediction accuracy even with less annotated data.

Overall, a notable contribution of this study is the development of a novel behavior-aware framework that moves beyond rigid classifications to capture the fluid and context-sensitive nature of driving behavior. This framework utilizes centrality measures and insights from traffic psychology to predict driver actions more effectively, enabling AVs to anticipate behaviors with a level of reasoning that reflects human-like intuition. Our model significantly outperforms state-of-the-art baseline models when tested on the NGSIM, HighD, RounD, ApolloScape, and MoCAD++ datasets. Remarkably, HiT maintains impressive performance even when trained on just 25\% of the dataset and with a much smaller number of model parameters. This demonstrates the model's robustness and adaptability across {various traffic scenarios}, including highways, roundabouts, campuses, and busy urban locales.

\section{Related Work}\label{Related work}

Trajectory prediction plays a central role in AD by enabling vehicles to anticipate the future positions of surrounding traffic agents. This capability is essential for making informed, real-time decisions to ensure both safety and efficiency. Over the years, researchers have developed and refined methodologies for trajectory prediction, which can be broadly categorized into three main approaches: physics-based, statistics-based, and deep learning-based methods.

\textbf{Physics-based Approaches.} Early work on trajectory prediction relied on physics-based models, such as vehicle kinematics and car-following models, which use the principles of physics and mechanics to predict future positions. These models leverage inputs such as velocity and steering angle to construct interpretable frameworks with low computational overhead \citep{schubert2008comparison, batz2009recognition}. Despite these advantages, their predictive accuracy is limited compared to state-of-the-art (SOTA) deep learning models.

\textbf{Statistics-based Approaches.} Statistical methods, including parametric and non-parametric models, aim to predict trajectories using predefined distributions. Examples include Gaussian processes, hidden Markov models, dynamic Bayesian networks, and support vector machines \citep{wang2021decision}. These methods offer improved accuracy over physics-based models because of their refined structures. For instance, \citet{xie2017vehicle} demonstrated significant performance gains by integrating a Kalman filter with a dynamic Bayesian network for vehicle trajectory prediction. However, these approaches suffer from key limitations: their accuracy lags behind deep learning models, and they rely on manual feature extraction tailored to specific scenarios, resulting in poor generalization.

\textbf{Deep Learning-based Approaches.}
Deep learning has emerged as a transformative approach for trajectory prediction.  \citet{alahi2016social} were among the first to model complex vehicle interactions using deep learning, employing convolutional neural networks (CNNs) to aggregate spatial information. More recently, attention mechanisms have been widely adopted for interaction modeling because of their ability to dynamically assign weights to the influence of different vehicles \citep{gupta2018social,liao2025minds}. Recognizing the interplay between spatial and temporal dynamics, \citet{chen2022intention} proposed the concepts of spatial and temporal interaction. Graph-based frameworks have also gained popularity for modeling unstructured road topologies \citep{liao2024mftraj}. Moreover, researchers have increasingly incorporated human cognitive factors into trajectory prediction. For instance, \citet{li2024context} highlighted the importance of the central field of vision for human drivers, whereas \citet{liao2024human} extended this concept by modeling the differential allocation of attention across various visual regions to capture essential spatio-temporal relationships.
Recent advancements explore the potential of large language models (LLMs) for trajectory prediction. Notable studies such as DiLu \citep{wen2024dilu} and CAVG \citep{liao2024gpt} leverage LLMs, including GPT-3.5 and GPT-4, for scene understanding and decision-making. Although these approaches achieve remarkable accuracy, they face significant challenges, including reliance on multimodal inputs, high computational costs, and latency issues that hinder real-time applications. Additionally, many deep learning models \citep{wang2023wsip, liao2024less} lack interpretability, operate as black boxes, and require large annotated datasets, limiting their scalability and adaptability to diverse driving scenarios.

In conclusion, the trajectory prediction domain faces an urgent need for innovative methodologies that address the critical challenges of interpretability, real-time processing, and scalability. We propose that integrating driver behavior—a crucial factor in driving decision-making—presents a promising direction for advancement. Variations in drivers' intrinsic behaviors can lead to different decisions, even under identical interaction patterns. However, current models predominantly focus on learning interaction patterns from datasets, neglecting the subjective and context-dependent nature of driver behavior. This limitation underscores the necessity of developing trajectory prediction models that can effectively account for such behavioral nuances. In a word, the same road scenario can induce different behaviors across drivers, shaped by their past experiences and driving styles, much like art evokes varied interpretations. Hence, effective trajectory prediction models should learn and represent the fine-grained nuances of driving behavior without manual intervention or discretization. By addressing this gap, future models can advance the field toward more interpretable, scalable, and behavior-aware trajectory prediction for autonomous vehicles.

\begin{figure}[t]
  \centering  \includegraphics[width=0.8\linewidth]{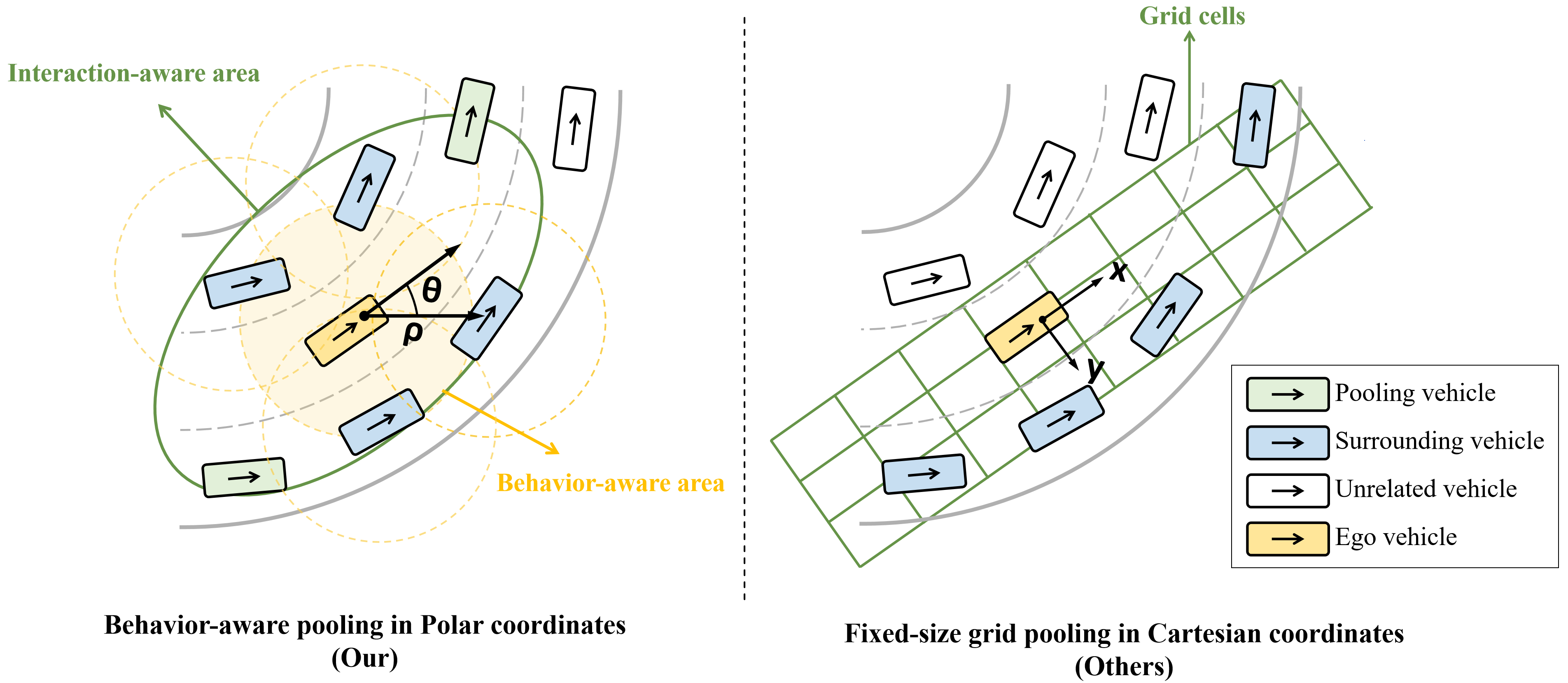} 
  \caption{Comparison of our proposed behavior-aware pooling mechanism using polar coordinates (left) versus the traditional fixed-size grid pooling mechanism using Cartesian coordinates (right). In the polar coordinate system, the ego vehicle (yellow) is centered at the origin, with surrounding vehicles (blue) mapped based on their relative radial distance and angle. This allows for a flexible, position-aware representation that better captures dynamic interactions, particularly in complex driving scenarios. The classical Cartesian grid-based approach (right) divides the environment into fixed-size cells, which may not accurately reflect the spatial relationships and dynamic behaviors in irregular or unstructured environments.}
  \label{polar} 
\end{figure}

\section{Problem Formulation}\label{Problem Formulation}

The goal of this study is to predict the trajectory of surrounding traffic agents within the AV’s sensing range in highly interactive driving scenarios. In line with standard practices \citep{huang2018apolloscape} and to better reflect real-world conditions, our model relies solely on local historical data of traffic agents within the AV's perceptual range for prediction. To simplify the analysis and avoid overly complex visualizations that may obscure comparisons with baseline methods, we focus on predicting the trajectory of a single representative vehicle, referred to as the ``ego vehicle'' or ``target vehicle''.
At each time step \(t\), the ego vehicle must predict the next \(t_f\) steps of the target vehicle's trajectory. Our model uses historical observations \(\bm{X}\) as inputs. These observations include the polar coordinates, velocity, and acceleration of the ego vehicle (subscript 0) and surrounding agents (subscript 1 to \(n\)) over a fixed time horizon \(t_h\). The model aims to predict a multimodal distribution of future trajectories for the ego vehicle, represented as \(P(\bm{Y}|\bm{X})\). Formally,
\begin{equation}\label{eq.1}
  \bm{X} = \bm{X}_{0:n}^{t-t_h:t} = \left\{(p_0^{t-t_h:t}, \mathbf{v}_0^{t-t_h:t}, \mathbf{a}_0^{t-t_h:t}), \ldots, (p_n^{t-t_h:t}, v_n^{t-t_h:t}, a_n^{t-t_h:t}) \right\}
\end{equation}
where \(p_0^{t-t_h:t}, \mathbf{v}_0^{t-t_h:t}, \mathbf{a}_0^{t-t_h:t}\) represent the polar position coordinates, velocity, and acceleration of the ego vehicle from time \(t-t_h\) to \(t\), respectively.

As illustrated in Fig. \ref{polar}, we depart from the conventional use of fixed grids or Cartesian coordinates and instead opt for a polar coordinate system. This choice is motivated by the need to better capture the relative spatial relationships in unstructured and dynamic driving environments. By anchoring the ego vehicle’s position at the origin of a static reference frame, the polar system allows for a more intuitive and flexible representation of the positions of surrounding vehicles. This enhances the model’s ability to accurately predict trajectories in diverse driving conditions, particularly in environments where the spatial relationships between agents are complex and non-linear. Specifically, at each time step \( t \), the position of the ego vehicle is anchored at the origin of a static reference frame. The position of the \( i \)th vehicle is then expressed as $p_{i}^{t}= \{\rho_{i}^{t}, \;\;\theta_{i}^{t}\}, \forall i\in[0,n]$, 
where \( \rho_{i}^{t} \) and \( \theta_{i}^{t} \) represent the radial distance and polar angle of the \( i \)th traffic agent, respectively. Correspondingly, the model's output is the future trajectory of the ego vehicle over the prediction interval $t_{f}$, which can be defined as $\bm{Y} = \bm{Y}_{0}^{t+1:t+t_{f}}= \left\{p_{0}^{t+1},p_{0}^{t+2},\ldots,p_{0}^{t+t_{f}-1},p_{0}^{t+t_{f}} \right\}$.

To address the inherent uncertainty and variability in prediction, our multimodal framework considers multiple potential maneuvers that the ego vehicle could perform, estimating the probability of each maneuver based on historical observations. See \textbf{Appendix \ref{Multimodal Probabilistic}} for details.

\section{Proposed Model}\label{Proposed Model}
\begin{figure}[t]
  \centering
  \includegraphics[width=0.8\linewidth]{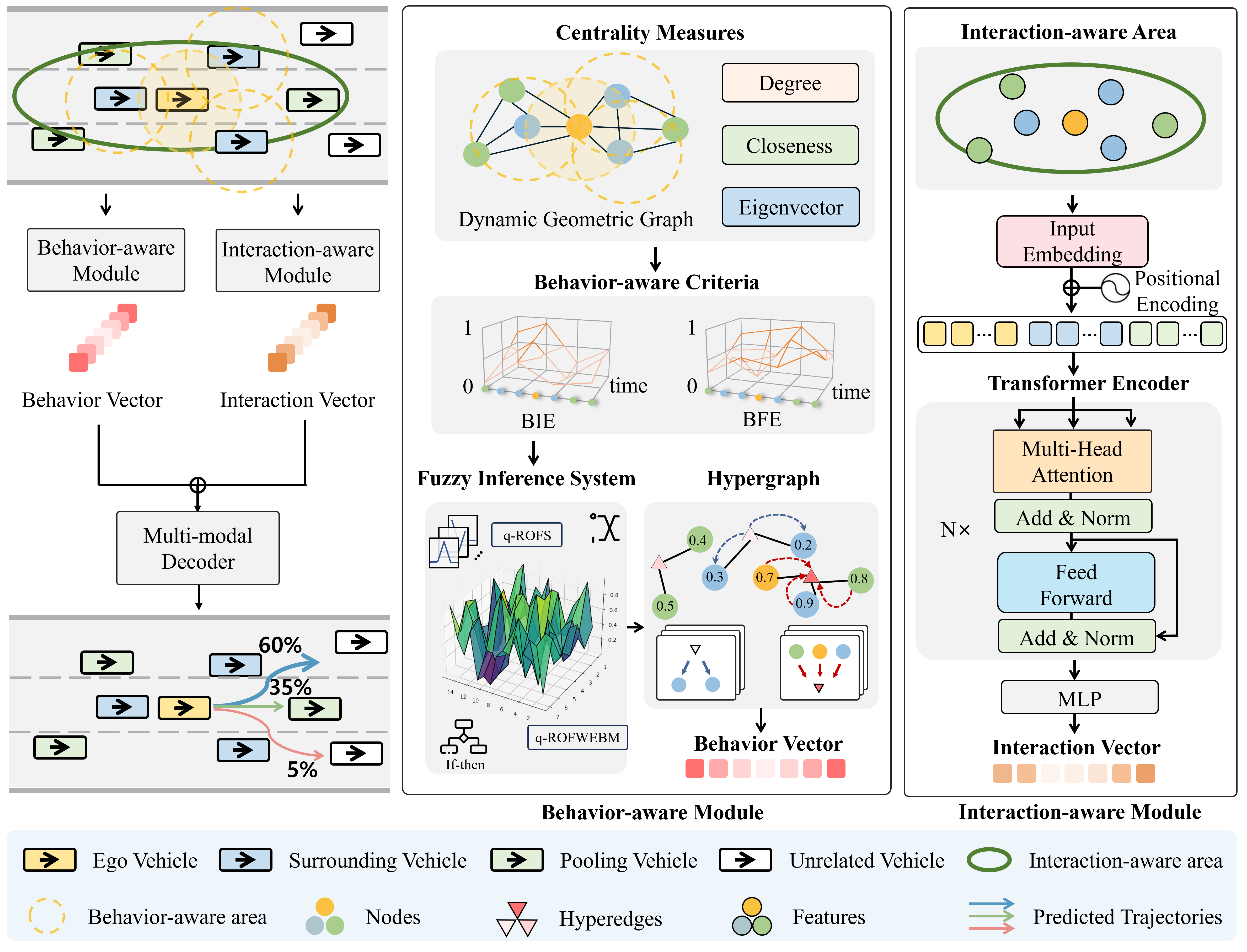} 
  \caption{Architecture of the proposed HiT model. The framework consists of three key components: the behavior-aware module, the interaction-aware module, and the multimodal decoder. The behavior-aware module leverages refined centrality measures to dynamically capture the evolving driving behavior of traffic agents. A novel behavior-aware criterion evaluates the intensity and patterns of these behaviors. Using these criteria, a fuzzy inference system quantifies driver aggressiveness, whereas a hypergraph neural network within the behavior encoder integrates and refines behavior features to generate behavior vectors $\mathbf{O}_{\text{beh}}$. Moreover, the interaction-aware module adopts a lightweight design with progressive positional encoding to generate spatio-temporal interaction vectors $\mathbf{O}_{\text{int}}$. Finally, the multimodal decoder fuses these vectors to predict multimodal trajectories $\bm{Y}_{0}^{t:t+t{f}}$.}
  \label{fig5} 
\end{figure}

\subsection{Overall Framework}
Fig. \ref{fig5} shows the architecture of our Human-Like Trajectory Prediction model (HiT). Built upon an encoder-decoder framework, HiT incorporates three key modules: the behavior-aware module, the interaction-aware module, and the multimodal decoder. The behavior-aware module is responsible for extracting the behavior vector $\mathbf{O}_{\text{beh}}$, which encapsulates the continuous driving behaviors of traffic agents. Simultaneously, the interaction-aware module captures spatial-temporal interactions among traffic agents, resulting in a high-level interaction vector $\mathbf{O}_{\text{int}}$. These vectors are then integrated within the multimodal decoder to generate diverse and accurate trajectory predictions.

\subsection{Behavior-aware Module}\label{Behavior-aware Module_0}
The behavior-aware module lies at the core of the HiT model, addressing the complex and dynamic nature of scenarios by offering a more flexible and adaptable approach to interpreting driver behavior. Unlike previous studies that rely on rigid, finite categories for driver behavior, our module embraces a continuous representation, allowing for a more nuanced understanding of driving actions. Inspired by the multi-policy decision-making (MPDM) framework for human drivers \citep{markkula2020defining} and enriched by traffic psychology insights, this module uses dynamic geometric graphs (DGGs) \citep{dall2002random} to model and assess driving behavior in real-time.

At the heart of this module is the use of DGGs to represent dynamic traffic scenarios, where the driving behaviors of agents are analyzed using refined centrality measures—specifically, dynamic degree centrality, dynamic closeness centrality, and eigenvector centrality. These centrality measures allow us to evaluate and quantify the influence and behavioral trends of each traffic agent within the network. Building on these insights, we introduce a novel behavior-aware criterion to assess the intensity and trajectory of driving behaviors. This criterion is processed through a Fuzzy Inference System (FIS), which applies fuzzification, fuzzy inference, and defuzzification to generate detailed scores reflecting the aggressiveness and other behavioral attributes of traffic agents. The output of this process is a high-order behavior vector, encoded by a behavior encoder, which provides a rich representation of behavior. This vector guides the model in generating human-like trajectory predictions, capturing the subtleties and variations in human driving preferences and patterns. By integrating these sophisticated methods, the behavior-aware module offers a comprehensive and precise depiction of driver behavior, enhancing the model's ability to predict realistic and context-sensitive trajectories.

\subsubsection{Dynamic Geometric Graphs}
DGGs are employed to represent the complex and evolving interactions between traffic agents in dynamic scenarios. At $t$, the traffic environment is modelled as a graph $G^{t}$, defined as ${G}^{t} = \{V^{t},{E}^{t}\}$. Here, $V^{t}=\{{v}_{0}^{t},{v}_{1}^{t}\ldots,{v}_{n}^t\}$ is the set of nodes where each node represents a vehicle, ${E^{t}} = \{{e_{0}^{t}},{e_{1}^{t}}\ldots,{e_{n}^{t}}\}$ is the set of undirected edges, where each edge ${e_{i}^{t}}$ signifies the potential influence between vehicles. The interaction between two vehicles $v_{i}$ and $v_{j}$ is considered significant if the shortest distance $d\left(v_{i}^{t}, v_{j}^{t}\right)$ between them is within a predefined threshold $r$. Formally, it can be defined as
${e_{i}^{t}}= \{({v_{i}^{t}}, {v_{j}^{t}}) \mid(j \in  {N}_{i}^{t})\}$
where $ {N}_{i}^{t}=\left\{v_{j}^{t} \in V^{t}\backslash\left\{v_{i}^{t}\right\}\mid d\left( v_{i}^{t}, v_{j}^{t}\right) \leq r, i \neq j\right\}$.

Correspondingly, the symmetrical adjacency matrix $A^{t}$ of $G^{t}$ can be given as follows:
\begin{equation}\label{eq.8}
A^{t}(i, j)= \begin{cases} 1 & \text { if } d\left(v_{i}^{t}, v_{j}^{t}\right)\leq{r}, i \neq j \\ 0 & \text {otherwise}\end{cases}
\end{equation}

\subsubsection{Centrality Measures}\label{Centrality Measures}
Centrality measures are critical tools in network analysis, quantifying the importance or influence of nodes within a graph.  In traffic networks, these measures facilitate a more human-like approach to trajectory prediction by capturing the dynamic interactions between traffic agents.
This subsection introduces the key centrality measures employed in this study.

\textbf{Dynamic Degree Centrality.}
In interactive traffic environments, a vehicle’s behavior is significantly shaped by its interactions with surrounding agents. Degree centrality, a fundamental concept in social networks, quantifies the number of direct connections a node has within a graph. When applied to traffic networks, it reflects the extent of a vehicle’s interactions with others, highlighting its potential to influence or be influenced by surrounding traffic agents of the AV.

\paragraph{(1) Classical Degree Centrality}

\begin{definition}\label{degree_1} \textit{Classical Degree Centrality:}
Given a DGG \( G^t = (V^t, E^t) \) at time \( t \), the degree centrality \( d_i^t \) of a node \( v_i^t \in V^t \) at time \( t \) is defined as the number of nodes directly connected to \( v_i^t \):
\begin{equation}
{d}_i^t = \sum_{j \in N_i^t} A^t(i, j)
\end{equation}
where \( N_i^t \) denotes the set of nodes adjacent to \( v_i^t \) at time \( t \).
\end{definition}

\begin{definition}\label{degree_2} \textit{Independence of Irrelevant Connections (IIC):}
A centrality measure \( c: {N}^t \rightarrow \mathbb{R}^n \) is said to exhibit IIC if, for all \( v_i^{t}, v_j^{t} \in V^{t} \setminus \{v_k^{t}, v_\ell^{t}\} \), the following implication holds:
\begin{equation}
c_i(G^{t}) \geq c_j(G^{t}) \Rightarrow c_i(G'^{t}) \geq c_j(G'^{t})
\end{equation}
\end{definition}

\begin{theorem} \label{Degree_th2}
    Classical degree centrality satisfies the IIC crieterion.
\end{theorem}

\begin{myproof}
Notably, we leave the proof for this theorem in \textbf{Appendix \ref{Proof_ICC_1}}.
\end{myproof}

\paragraph{(2) Proposed Dynamic Degree Centrality} 
Classical degree centrality, as established in Theorem \ref{Degree_th2}, is inherently limited in its application to dynamic traffic scenarios. It focuses solely on the number of direct connections (or neighbors) a node has, without considering the qualitative nature of these connections or their broader influence within DGGs. This narrow focus can lead to delayed or inaccurate measurements in rapidly changing environments. 

For example, at time \( t \), consider a scenario where an interaction exists between traffic agents \( v_i \) and \( v_j \), such that \( d(v_i^t, v_j^t) \leq r \). If vehicle \( v_j \) suddenly changes lanes or accelerates, classical degree centrality may fail to promptly capture the real-time impact of this action on the entire traffic network. The measure only adjusts at a later time \( t_k \) (\( t < t_k \)), once the physical distance between vehicles has significantly changed (e.g., \( d(v_i^{t_k}, v_j^{t_k}) > r \)), potentially missing the critical response window needed for optimal decision-making. In more complex multi-agent traffic scenarios, such as multi-lane merging or roundabouts, where multiple vehicles interact simultaneously, classical degree centrality may also underestimate the significance of certain critical vehicles—especially those newly introduced into the traffic flow. These vehicles, despite forming new connections, may exhibit changes in speed or acceleration that profoundly affect overall traffic dynamics. However, classical degree centrality, which relies on counting direct connections, fails to differentiate between nodes connected to central versus peripheral nodes. This shortcoming can delay the system's response at crucial moments.

To address these limitations, we propose \textbf{Dynamic Degree Centrality}, a novel measure specifically tailored for traffic scenarios. Unlike classical degree centrality, this measure does not strictly adhere to the IIC criterion, allowing it to better capture the real-time, dynamic nature of interactions within a traffic network. Dynamic degree centrality assesses not only the direct connections of a node but also the influence of the nodes to which it is connected (indirect connections). This approach offers a more nuanced and responsive measure of a vehicle's influence within the traffic network, effectively capturing the immediate effects of dynamic changes such as lane shifts, sudden accelerations, or the merging of traffic flows. The dynamic degree centrality is defined using a kernel-weighted multimodal adjacency matrix and a corresponding Laplacian matrix, which together account for varying strengths of interactions based on factors like distance, speed, and acceleration.

\begin{definition} \textit{Kernel-weighted Multimodal Adjacency Matrix:}
Given a DGG \( G^t = (V^t, E^t) \) at time \( t \), for any pair of vehicles \( v_i^t \) and \( v_j^t \), the elements of the kernel-weighted multimodal adjacency matrix \( \bar{A}^t = \left(\bar{a}_{ij}^t\right) \) can be mathematically defined as follows:
\begin{equation}
\bar{a}_{ij}^t = \exp\left(-\sum_{k} \frac{|x_{i,k}^t - x_{j,k}^t|^2}{2\bar{\sigma}_k^2}\right)
\end{equation}
where \( x_{i,k}^t \) represents the \( k \)-th dimensional feature of the vehicle \( v_i^t \) (including radial distance \( \rho_i^t \), polar angle \( \theta_i^t \), velocity \( v_i^t \), and acceleration \({a}_i^t \)) and \(\bar{\sigma}_k \) is the standard deviation for the \( k \)th dimension, controlling the influence weight of that dimension on the adjacency matrix elements. Moreover, the corresponding kernel-weighted multimodal Laplacian Matrix \( \bar{L}^t \) is defined as $\bar{L}^t = \bar{D}^t - \bar{A}^t$, where traditional degree matrix \( \bar{D}^t_i \) is the a diagonal matrix with elements \( \ell_{ii}^t = \sum_{j \in N_i^t} \bar{a}_{ij}^t \), and \( N_i^t \) denotes the set of nodes directly connected to node \( v_i^t \) at time $t$.
\end{definition}

\begin{definition} \textit{Dynamic Degree Centrality:}
Given a DGG \( G^t = (V^t, E^t) \) at time \( t \), the dynamic degree centrality \( \mathbf{x}^t(\varepsilon): V^t \rightarrow \mathbb{R}^n \) is defined by the equation $
(I + \varepsilon L^t) \mathbf{x}^t(\varepsilon) = \mathbf{d}^t$, where \( I \) is the identity matrix, \( \mathbf{d}^t = (d_1^t, d_2^t, \ldots, d_n^t)^T \) represents the traditional degree centrality vector at time \( t \), and \( \varepsilon > 0 \) is a parameter that controls the influence of indirect connections. Formally, the dynamic degree centrality \( \mathcal{J}_{i}^{t}(D) \) for a node \( v_i^t \) is expressed as follows:
\begin{equation}
\mathcal{J}_{i}^{t}(D) = x_i^t(\varepsilon) + \varepsilon \sum_{v_j^t \in N_i^t} \bar{a}_{ij}^t \left[x_i^t(\varepsilon) - x_j^t(\varepsilon)\right]
\end{equation}
where \( x_i^t(\varepsilon) \) denotes the degree centrality for \( v_i^t \), considering the effects of indirect interaction. 
\end{definition}

\begin{theorem}
Dynamic degree centrality violates the IIC criterion for any \( \varepsilon > 0 \).
\end{theorem}
\begin{myproof}
The proof of this theorem is provided in \textbf{Appendix \ref{Proof_ICC_2}}.
\end{myproof}

Expanding upon the previously established theorem and theoretical principles, the proposed dynamic degree centrality overcomes the limitations of the IIC by integrating the effects of indirect connections. This enhancement provides a more detailed and timely evaluation of the interactions among traffic agents in dynamic scenarios. This centrality metric is especially adept at capturing the immediate effects of abrupt changes, such as lane changes or sudden acceleration, thus providing the model with a more insightful and context-aware understanding of dynamic traffic interactions.

\textbf{Dynamic Closeness Centrality.}
Closeness centrality measures the ability of a node to efficiently connect with other nodes in a graph. In the context of traffic graphs, this metric reflects the influence of an agent, where greater accessibility to its peers increases its likelihood to facilitate interactions. To better capture the behavior of traffic agents, we use closeness centrality to evaluate the proximity of each vehicle to its neighbors. A vehicle positioned closer to its surroundings has a higher probability of interacting with nearby vehicles, embodying both convenience and accessibility for interactions. This aligns with the notion that vehicles with higher accessibility are more effective in influencing and inducing interactions within the traffic network \citep{ahmed2015sentiment}.

However, classical closeness centrality assumes fixed connection strengths between nodes, which is effective in stable networks like social graphs but inadequate for dynamic and rapidly changing traffic scenarios. In traffic environments, factors such as relative position, speed, acceleration, and steering angle continuously change in real time, significantly impacting path planning and decision-making. Conventional metrics, which rely on static topological distances, struggle to model these complex and dynamic interactions effectively. To overcome this limitation, we propose a novel \textbf{dynamic closeness centrality} framework tailored for traffic scenarios. Technically, we redefine the distance metric within a dynamic geometric graph using polar coordinates, with the ego vehicle as the origin. This approach represents the positions of other traffic agents in terms of radial distance and angular displacement, providing an intuitive spatial representation relative to the ego vehicle. Beyond spatial relationships, we integrate dynamic variables such as speed and acceleration into the distance metric, capturing movement trends and enabling a more comprehensive representation of interactions. Additionally, we also implement a dynamic weighting factor that adjusts the interaction strength between nodes in response to real-time variations in traffic conditions. This enhancement ensures that the closeness centrality measure remains highly responsive to evolving traffic dynamics, improving the model’s adaptability to sudden changes in the complex traffic environment.

Specifically, the calculation of the closeness centrality of a node in the graph is based on the shortest paths between the given vehicle and all other surrounding agents. This is achieved by taking the inverse of the sum of the distances between the node and its surrounding vehicles. Therefore, the dynamic closeness centrality $ {J}_{i}^{t}(C)$ for the $i$th traffic agents in the DGG can be given as follows: 
\begin{equation}\label{eq.12}
\mathcal{J}_{i}^{t}(C)=\frac{\left| \mathcal{N}_{i}^{t}\right|}{\sum_{\forall v_{j}^{t} \in \mathcal{N}^{t}_{i}}\bar{w}_{ij}^{t}\bar{d}\left(v_{i}^{t}, v_{j}^{t}\right)}
\end{equation}
where $\left| \mathcal{N}{i}^{t}\right|$ denotes the total elements in $\mathcal{N}{i}^{t}$. $\bar{w}_{ij}^{t}$ is the dynamic weighting factor, which can be defined as $\bar{w}_{ij}^{t} = \exp\left(-\left[\gamma_1 |\rho_i^t - \rho_j^t| + \gamma_2 |\theta_i^t - \theta_j^t| + \gamma_3 |v_i^t - v_j^t| + \gamma_4 |a_i^t - a_j^t|\right]\right)$, with $\gamma_1, \gamma_2, \gamma_3, \gamma_4$ are parameters that regulate the influence of each dynamic variable on the weighting.  Moreover, the distance metric \( \bar{d}\left(v_{i}^{t}, v_{j}^{t}\right) \) is formally formulated as follows:
\begin{equation}
\bar{d}\left(v_{i}^{t}, v_{j}^{t}\right) = \sqrt{\eta_1 (\rho_i^t - \rho_j^t)^2 + \eta_2 (\theta_i^t - \theta_j^t)^2 + \eta_3 \|\mathbf{v}_i^t - \mathbf{v}_j^t\|^2 + \eta_4 \|\mathbf{a}_i^t - \mathbf{a}_j^t\|^2} 
\end{equation}

 Here, \( \eta_1, \eta_2, \eta_3, \eta_4 \) are all the weighting parameters that control the contribution of each factor to the overall distance. 
The optimized closeness centrality measure evolves from a static metric into one that can adapt to dynamic changes. It is worth noting that a higher value of closeness centrality indicates a more central position for the node. In other words, as the vehicle gets closer to the surrounding agents, its dynamic closeness centrality increases.

\textbf{Eigenvector Centrality.}
Eigenvector centrality is a measure of the importance of a node in a network, based on the principle that a node is considered important if it is connected to other important nodes. In the context of understanding driver behavior, the eigenvector centrality measure of a vehicle takes into account both the number of interactions with its surrounding vehicles and the influence of the vehicles with which it interacts. 
The eigenvector centrality measure of a vehicle takes into account both the number of connections the vehicle has and the degree of influence of the vehicles it is connected to. This allows us to identify the most influential vehicles in a traffic scenario and understand how their actions and decisions may affect the behavior of other drivers. Formally,
\begin{equation}\label{eq.13}
\mathcal{J}_{i}^{t}(E)=\frac{ \sum_{\forall v_{j}^{t} \in \mathcal{N}^{t}_{i}}d\left(v_{i}^{t}, v_{j}^{t}\right)}{\lambda}
\end{equation}
where $\lambda$ is the eigenvalue. In addition, the Perron-Frobenius theorem states that for a non-negative matrix (such as the adjacency matrix in our case), there exists a positive eigenvector solution for the greatest eigenvalue of the matrix \citep{pillai2005perron}. This implies that the eigenvector associated with the dominant eigenvalue of the adjacency matrix can be employed to determine the eigenvector centrality measure of the nodes in the graph. This approach is advantageous as it facilitates the identification of the most crucial nodes in the DGGs, based on their connections to other vital nodes, thereby providing valuable insights into the driving behaviors of traffic agents. Moreover, the utilization of the Perron-Frobenius theorem in this context ensures that the eigenvector centrality measure is well-defined and unique, thereby guaranteeing consistent and reliable outcomes.

\subsubsection{Behavior-aware Criteria}

Centrality measures, such as Dynamic Degree Centrality, Dynamic Closeness Centrality, and Eigenvector Centrality, provide a detailed analysis of the interactions and influence of each vehicle within the network. However, to effectively utilize these insights for real-time trajectory prediction, we need a mechanism to interpret and quantify these behaviors dynamically. This is where the \textbf{Behavior-aware Criteria} come into play.

\textbf{Behavior Intensity Estimate (BIE).} 
The BIE measures the rate of change in the centrality values over time, reflecting how drastically a vehicle's behavior is deviating from the norm. This criterion is particularly useful in identifying sudden maneuvers such as sharp lane changes or rapid acceleration. Mathematically, it can be computed as follows:
\begin{equation}\label{eq.14}
{\mathcal{I}}_{i}^{t}=\left[\left|\frac{\partial \mathcal{J}^{t}_{i}(D)}{\partial t}\right|,\left|\frac{\partial \mathcal{J}^{t}_{i}(C)}{\partial t }\right|,\left|\frac{\partial \mathcal{J}^{t}_{i}(E)}{\partial t }\right|\right]^{T}
\end{equation}
where $\left| \cdot  \right|$ is the absolute value operator ensuring we capture both increases and decreases in centrality, which reflects changes in behavior intensity. A higher BIE indicates significant changes in behavior.

\textbf{Behavior Fluctuation Estimate (BFE).}
Although BIE captures the intensity of a behavior, the Behavior Fluctuation Estimate (BFE) quantifies the consistency or variability of that behavior over time. A stable driving pattern will have a low BFE, indicating consistent behavior, whereas erratic driving will result in a high BFE, signaling significant fluctuations.

The BFE is calculated by taking the second derivative of the centrality measures, which provides insights into the acceleration or deceleration of behavioral intensity:
\begin{flalign}\label{eq.15}
{{\mathcal{F}}}_{i}^{t}  =\left|\frac{\partial\mathcal{{{\mathcal{I}}}}_{i}^{t}}{\partial t}\right|  
     = \left[\left|\frac{\partial^{2} \mathcal{J}_{i}^{t}(D)}{\partial^{2} t}\right|,\left|\frac{\partial^{2} \mathcal{J}_{i}^{t}(C)}{\partial^{2} t }\right|, \left|\frac{\partial^{2} \mathcal{J}_{i}^{t}(E)}{\partial^{2} t }\right|\right]^{T} 
\end{flalign}
This measure is particularly useful for detecting sudden, unpredictable changes in behavior, such as a vehicle that suddenly swerves or brakes sharply.

These criteria provide a quantifiable approach to understanding how vehicles behave within a traffic network, offering real-time insights into potential risks. In the following sections, we denote the sets of BIE and BFE for all observed agents across the entire historical observations as ${\mathcal{J}} = \{\mathcal{I}, \mathcal{F}\}$, with \( \mathcal{I}= {\mathcal{I}}_{0:n}^{t-t_h:t}\), and \(\mathcal{F}= {\mathcal{F}}_{0:n}^{t-t_h:t}\). For non-differentiable time points, we use zero padding.

\subsubsection{Fuzzy Inference Processing}
Building on the Behavior-aware Criteria, we propose a novel Fuzzy Inference System (FIS) that systematically translates these criteria into a comprehensive score indicating the level of driving aggressiveness. Our approach follows the standard fuzzy logic framework \citep{mendel1995fuzzy}, consisting of three key steps: fuzzification, fuzzy inference, and defuzzification. This process provides a more nuanced and interpretable assessment of aggressiveness.  The detailed process of fuzzy inference is outlined below.

\textbf{Step-1: Fuzzification.}
The fuzzification process is responsible for mapping the input data, specifically the behavior criteria BIE and BFE, into predefined fuzzy sets by calculating their membership degrees using membership functions. We utilize triangular membership functions to define distinct fuzzy sets for each criterion. To avoid the subjectivity often associated with traditional expert-based methods, we adopt a data-driven approach for defining the fuzzy set boundaries of BIE and BFE.

\begin{figure}[t]
  \centering
\includegraphics[width=1\linewidth]{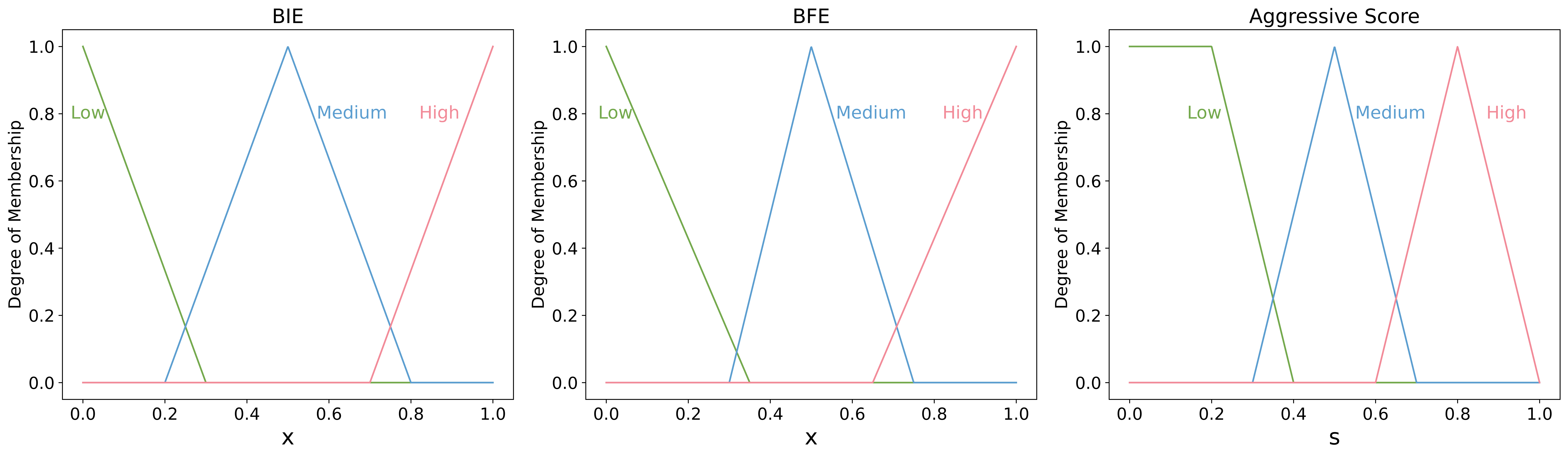} 
  \caption{Membership functions of (a) BIE, (b) BFE, and (c) Aggressive score. ``Low'', ``Medium'', and ``High'' denote the Low-level, Medium-level, and High-level categories for the fuzzy sets, respectively.}
  \label{Membership} 
\end{figure}

Initially,  both the BIE and the BFE for $v_{i}^{t}$ at time $t$ undergo standard normalization preprocessing, denoted as $\mathcal{\bar{I}}_{i}^{t}, \in [0,1] $, $\mathcal{\bar{F}}_{i}^{t} \in [0,1]$ respectively, followed by multiple iterations of K-means \citep{hamerly2003learning} clustering analysis. By evaluating the Sum of Squared Errors (SSE) across different cluster numbers, we identify the ``elbow point'' to determine the optimal fuzzy set partitioning. The resulting membership functions for the behavior-aware criteria and output, depicted in Fig. \ref{Membership}, categorize BIE and BFE into three fuzzy sets: ``Low'' (L), ``Medium'' (M), and ``High'' (H), using the cluster centroids as the boundary points. Notably, the output is the aggressive score, which indicates the level of aggressiveness of the drivers.

The  Low (L)/Medium (M)/High (H)-level fuzzy sets for BIE are represented as follows:
\begin{equation}
  A_{\text{L,M,H}} = \{ x_{{L,M,H}}^{\text{BIE}} \in \mathcal{\bar{I}} \mid \mu_{\text{L,M,H}}^{\text{BIE}}(x_{{L,M,H}}^{\text{BIE}}) \}, \quad x_{\text{L}}^{\text{BIE}} \in [0, 0.3], \,
  x_{\text{M}}^{\text{BIE}} \in [0.2, 0.8], \,
  x_{\text{H}}^{\text{BIE}} \in [0.7, 1],\,
  \end{equation}

Similarly, the Low/Medium/High-level fuzzy sets for BFE are defined as follows:
 \begin{equation}
  B_{\text{L,M,H}} = \{ x_{{L,M,H}}^{\text{BFE}} \in \mathcal{\bar{F}} \mid \mu_{\text{L,M,H}}^{\text{BFE}}(x_{{L,M,H}}^{\text{BIE}}) \}, \quad x_{{L}}^{\text{BFE}} \in [0, 0.35], \, 
  x_{{M}}^{\text{BFE}} \in [0.3, 0.75],\, 
  x_{{H}}^{\text{BFE}} \in [0.65, 1],
  \end{equation}

Additionally, the Low/Medium/High-level fuzzy sets for aggressive scores are defined:
 \begin{equation}
  C_{\text{L,M,H}} = \{ s_{{L,M,H}} \in \mathcal{\bar{F}} \mid \mu_{\text{L,M,H}}^{\text{Score}}(s_{{L,M,H}}) \}, \quad s_{{L}} \in [0, 0.4], \, 
  s_{{M}} \in [0.3, 0.7],\, 
  s_{{H}} \in [0.6, 1],
  \end{equation}

Each fuzzy set is defined by a membership function $\mu(x)$, which maps specific BIE and BFE values to their degrees of membership. The definitions of the Low, Medium, and High membership functions for BIE, BFE, and aggressiveness scores are detailed in \textbf{Appendix \ref{FIS_1}}.

After computing the membership degrees for each fuzzy set, we further capture the uncertainty by calculating the non-membership degree \( \nu(x) \) to form fuzzy number pairs. To simplify, we represent the fuzzy number pair for BIE and BFE as \( Q = \{Q_{\text{BIE}}, Q_{\text{BFE}}\} = \langle \mu(x), \nu(x) \rangle, x \in \{\mathcal{\bar{I}}, \mathcal{\bar{F}}\} \). The non-membership degree is computed using the q-Rung Ortho-pair Fuzzy Set (q-ROFS) method \citep{peng2021review}, which can be defined as follows:
\begin{equation}
\nu(x) = \left(1 - \mu(x)^q\right)^{\frac{1}{q}}
\end{equation}
where \( q \geq 1 \) is a control parameter that ensures that the computed subordination and non-subordination degrees satisfy the following condition:
\begin{equation}
\mu(x)^q + \nu(x)^q \leq 1
\end{equation}

\begin{table}[t]
  \centering
  \caption{Fuzzy rules for classifying driver aggressiveness levels.   ``H'' indicates high aggressiveness, ``M'' indicates medium aggressiveness, and ``L'' indicates low aggressiveness.}
   \resizebox{0.4\linewidth}{!}{
    \begin{tabular}{c|c|c|c|c}
    \hline
    Index & BIE   & Operator & BFE   & Output \\
    \hline
    1     & H     & And   & H     & H \\
    2     & H     & And   & M     & H \\
    3     & H     & And   & L     & H \\
    4     & M     & And   & H     & H \\
    5     & M     & And   & M     & M \\
    6     & M     & And   & L     & M \\
    7     & L     & And   & H     & M \\
    8     & L     & And   & M     & L \\
    9     & L     & And   & L     & L \\
    \hline
    \end{tabular}%
    }
  \label{rule}%
\end{table}%

\textbf{Step-2: Fuzzy Inference.}
As depicted in Table \ref{rule}, we meticulously establish a series of fuzzy rules that indicate how the FIS should assess the input behavior-aware criteria. These fuzzy rules adhere to the traditional ``If-Then'' format, where the antecedent is formed from the fuzzy sets of BIE and BFE, and the consequent assigns the fuzzy set corresponding to the output aggressive scores. The fuzzy reasoning process is carried out by employing these rules to the input criteria and their fuzzy number pairs within the pertinent fuzzy sets, ultimately producing a fuzzy outcome.

For instance, if a rule's antecedent is defined by the fuzzy number pairs \( Q_{\text{BIE}} = \langle \mu_{\text{BIE}}, \nu_{\text{BIE}} \rangle \), \( Q_{\text{BFE}} = \langle \mu_{\text{BFE}}, \nu_{\text{BFE}} \rangle \), and the resulting aggressive scores is represented by \( Q_{\text{Agg}} = \langle \mu_{\text{Agg}}, \nu_{\text{Agg}} \rangle \), a standard fuzzy rule can be expressed as: \textit{If BIE is High and BFE is Medium, then the Aggressiveness Level is High}.
Each fuzzy rule processes the input fuzzy number pairs and generates an output fuzzy number pair that reflects the combined influence of the input conditions. Here, \( \mu_{\text{BIE}} \), \( \mu_{\text{BFE}} \), and \( \mu_{\text{Agg}} \) represent the degrees of BIE, BFE, and the resulting aggressive scores, respectively. Similarly, \( \nu_{\text{BIE}} \), \( \nu_{\text{BFE}} \), and \( \nu_{\text{Agg}} \) correspond to the non-membership degrees of BIE, BFE, and the resulting aggressive scores.

\textbf{Step-3: Defuzzification.}
This process aims to denazify the aggregated fuzzy result \( Q_{\text{Agg}} = \langle \mu_{\text{Agg}}, 
u_{\text{Agg}} \rangle \) into a precise value that reflects the overall aggressiveness level. In this study, we utilize the \textbf{q-rung ortho-pair fuzzy weighted Einstein Bonferroni mean (q-ROFWEBM) operator}, a choice driven by its ability to adeptly manage the complexities in traffic scenarios.

The q-ROFWEBM operator is particularly suited for handling the nonlinear relationships inherent in dynamic driving behaviors. Its use of Einstein addition and multiplication techniques allows it to process these complex interactions effectively, ensuring that the defuzzified output accurately represents the intricate relationships within the data. Additionally, the q-ROFWEBM operator offers a refined balance between flexibility and precision. As an enhancement of the q-rung ortho-pair fuzzy weighted Archimedean BM (q-ROFWABM) operator, it provides a more sophisticated approach to information aggregation. This operator is also designed for computational efficiency, delivering rapid results without compromising accuracy, making it ideal for real-world applications. Finally, the smoothness of the output produced by the q-ROFWEBM operator is a significant advantage, as it avoids abrupt changes and ensures that the results are consistent and predictable.

To validate the robustness of the q-ROFWEBM operator in this context, we first establish its formulation based on the fundamental properties of the q-rung ortho-pair fuzzy Archimedean BM (q-ROFABM) operator. This foundation ensures that the operator adheres to the mathematical principles governing q-rung ortho-pair fuzzy numbers. Subsequently, we detail the defuzzification process, ensuring that the integrated results remain consistent within this framework. The following definitions and theorems formally establish the operator's properties, preserving the integrity of the defuzzification process whereas maintaining computational validity.

\begin{definition} \label{q-ROFABM} \textit{q-ROFABM Operator:}
Let \( Q_s = \langle u_c, v_\epsilon \rangle \), where \( \varepsilon = 1, 2, \dots, m \), as a collection of q-rung ortho-pair fuzzy numbers (q-ROFNs), with \( o, h \geq 0 \) and \( q \geq 1 \). The operator \( \mathrm{q}-\operatorname{ROFABM}: \Omega^m \rightarrow \Omega \) is defined as the q-rung ortho-pair Archimedean BM (q-ROFABM) operator if it satisfies the following condition:
\begin{equation}
\mathrm{q}-\operatorname{ROFABM}^{o,h}(Q_1, Q_2, \dots, Q_m) = \left(\frac{1}{m(m-1)} \sum_{\substack{i, j=1 \\ i \neq j}}^m Q_i^o \otimes_A Q_j^h \right)^{\frac{1}{o+h}}
\end{equation}
where \( \Omega \) denotes the set of all q-rung ortho-pair fuzzy numbers.
\end{definition}

\begin{definition} \label{q-ROFWABM}\textit{q-ROFWABM Operator:}
Let \( Q_{\varepsilon} = \langle u_{\varepsilon}, v_{\varepsilon} \rangle \) for \(\varepsilon = 1, 2, \dots, m\) denote a collection of \( Q \)-order pair fuzzy numbers. Assume that \( o, h \geq 0 \) and \( q \geq 1 \). The operator q-order pair fuzzy weighted Archimedean BM (q-ROFWABM) operator is defined as a mapping q-ROFWABM: \(\Omega^m \rightarrow \Omega\), where
\begin{equation}
\mathrm{q}-\operatorname{ROFWABM}^{o, h}\left(Q_1, Q_2, \dots, Q_m\right) = \left(\frac{1}{m(m-1)} \sum_{\substack{i, j=1 \\ i \neq j}}^m \left(m w_i Q_i\right)^o \otimes_A \left(m w_j Q_j\right)^h \right)^{\frac{1}{o+h}}
\end{equation}

Here, \(\Omega\) represents the entire set of \( Q \)-order pair fuzzy numbers, \( w_{i} = \left(w_1, w_2, \dots, w_m\right)^{\mathrm{T}} \) and \( w_{j} = \left(\bar{w}_1, \bar{w}_2, \dots, \bar{w}_m\right)^{\mathrm{T}} \) are both the weight vector associated with \( \left(Q_1, Q_2, \dots, Q_m\right) \), satisfying \( w_{\varepsilon} \in [0,1] \) and \( \sum_{\varepsilon=1}^m w_{\varepsilon} = 1 \). Under these conditions, the q-ROFWABM operator is referred to as the \( q \)-order pair fuzzy weighted Archimedean BM operator.
\end{definition}

\begin{theorem} \label{q-ROFABM_th1}
Let \( Q_{\varepsilon} = \langle u_{\varepsilon}, v_{\varepsilon} \rangle \) (for \( \varepsilon = 1, 2, \dots, m \)) be a set of q-rung ortho-pair fuzzy numbers, with \( q \geq 1 \). Then the integrated result of the q-ROFABM operator, denoted as \( \mathrm{q}-\operatorname{ROFABM}^{o, h}(Q_1, Q_2, \dots, Q_m) \), remains a q-rung ortho-pair fuzzy number and satisfies the expression:
\begin{equation}
\begin{aligned}
&\mathrm{q}-\operatorname{ROFABM}^{o, h}(Q_1, Q_2, \dots, Q_m) \\
&= \left\langle g^{-1}\left(\frac{1}{o+h} g\left(f^{-1}\left(\frac{1}{m(m-1)}\sum_{\substack{i,j=1 \\ i \neq j}}^m f\left(g^{-1}\left(o g(u_i) + h g(u_j)\right)\right)\right)\right)\right),\right. \\
&\quad \left. f^{-1}\left(\frac{1}{o+h} f\left(g^{-1}\left(\frac{1}{m(m-1)}\sum_{\substack{i,j=1 \\ i \neq j}}^m g\left(f^{-1}\left(o f(v_i) + h f(v_j)\right)\right)\right)\right)\right)\right\rangle
\end{aligned}
\end{equation}
where \( g^{-1} \) and \( f^{-1} \) denote the inverse functions of \( g \) and \( f \), respectively.
\end{theorem}

\begin{myproof}
From Definition \ref{q-ROFABM}, the q-ROFABM operator is defined as:
\begin{equation}
  \mathrm{q}-\operatorname{ROFABM}^{o,h}(Q_1, Q_2, \dots, Q_m) = \left(\frac{1}{m(m-1)} \sum_{\substack{i,j=1 \\ i \neq j}}^m Q_i^o \otimes_A Q_j^h \right)^{\frac{1}{o+h}}  
\end{equation}

Here, \( Q_i = \langle u_i, v_i \rangle \) and \( Q_j = \langle u_j, v_j \rangle \) are q-rung ortho-pair fuzzy numbers, and the operation \( \otimes_A \) represents the product operation of the fuzzy Archimedean operator.

We begin by computing \( Q_i^o \otimes_A Q_j^h \). Given \( Q_i^o = \langle u_i^o, v_i^o \rangle \) and \( Q_j^h = \langle u_j^h, v_j^h \rangle \), we express them as follows:
\begin{equation}
Q_i^o = \left\langle g^{-1}\left(o g(u_i)\right), f^{-1}\left(o f(v_i)\right)\right\rangle
\end{equation}
\begin{equation}
Q_j^h = \left\langle g^{-1}\left(h g(u_j)\right), f^{-1}\left(h f(v_j)\right)\right\rangle
\end{equation}

Next, we combine \( u_i \) and \( u_j \), starting by calculating the inner term of \( g^{-1} \):
\begin{equation}
u_{ij}^o = g\left(g^{-1}\left(o g(u_i)\right)\right) + g\left(g^{-1}\left(h g(u_j)\right)\right)
\end{equation}
\begin{equation}
u_{ij}^o = o \cdot g(u_i) + h \cdot g(u_j)
\end{equation}
such that,
\begin{equation}
u_{ij} = g^{-1}\left(g\left(g^{-1}\left(o g(u_i)\right)\right) + g\left(g^{-1}\left(h g(u_j)\right)\right)\right) = g^{-1}\left(o \cdot g(u_i) + h \cdot g(u_j)\right)
\end{equation}

Similarly, for \( v_i \) and \( v_j \), we proceed as follows:
\begin{equation}
v_{ij}^o = f\left(f^{-1}\left(o f(v_i)\right)\right) + f\left(f^{-1}\left(h f(v_j)\right)\right)
\end{equation}
% This can be simplified as follows:
\begin{equation}
v_{ij}^o = o \cdot f(v_i) + h \cdot f(v_j)
\end{equation}

Substituting this result back into \( f^{-1} \), we get the following equation:
\begin{equation}
v_{ij} = f^{-1}\left(v_{ij}^o\right) = f^{-1}\left(o \cdot f(v_i) + h \cdot f(v_j)\right)
\end{equation}
Thus, the result of \( Q_i^o \otimes_A Q_j^h \) is:
\begin{equation}
Q_i^o \otimes_A Q_j^h = \left\langle u_{ij}, v_{ij} \right\rangle = \left\langle g^{-1}\left(o \cdot g(u_i) + h \cdot g(u_j)\right), f^{-1}\left(o \cdot f(v_i) + h \cdot f(v_j)\right)\right\rangle
\end{equation}
Then, we sum and average all \( Q_i^o \otimes_A Q_j^h \) over all possible \( (i, j) \) combinations.
the sum and average of all \( u_{ij} \) is represented as follows:
\begin{equation}
u' = \frac{1}{m(m-1)} \sum_{\substack{i,j=1 \\ i \neq j}}^m g^{-1}\left(o \cdot g(u_i) + h \cdot g(u_j)\right)
\end{equation}
Given the nonlinearity of the \( g^{-1} \) function, we apply it after summing:
\begin{equation}
u' = g^{-1}\left(\frac{1}{m(m-1
)} \sum_{\substack{i,j=1 \\ i \neq j}}^m \left(o \cdot g(u_i) + h \cdot g(u_j)\right)\right)
\end{equation}

Similarly, for \( v_{ij} \):
\begin{equation}
v' = f^{-1}\left(\frac{1}{m(m-1)} \sum_{\substack{i,j=1 \\ i \neq j}}^m \left(o \cdot f(v_i) + h \cdot f(v_j)\right)\right)
\end{equation}

Thus, the result for the entire set \( \frac{1}{m(m-1)} \sum_{\substack{i,j=1 \\ i \neq j}}^m Q_i^o \otimes_A Q_j^h \) is represented as follows:
\begin{equation}
\begin{aligned}
 \frac{1}{m(m-1)} \sum_{\substack{i, j=1 
i \neq 1}}^m Q_i^o \otimes_A Q_j^h =
\left\langle u', v' \right\rangle&= \left\langle f^{-1}\left(\frac{1}{m(m-1)}\left(\sum_{\substack{i, j=1 \\
i \neq j}}^m f\left(g^{-1}\left(o \cdot g\left(u_i\right)+\operatorname{h\cdot g}\left(u_j\right)\right)\right)\right)\right),\right. \\
& \left.g^{-1}\left(\frac{1}{m(m-1)}\left(\sum_{\substack{i, j=1 \\
l \neq j}}^m g\left(f^{-1}\left(\operatorname{o \cdot h}\left(v_i\right)+h \cdot f\left(v_j\right)\right)\right)\right)\right) \right\rangle
\end{aligned}
\end{equation}

Ultimately, the power function transformation is applied, and the previously calculated expressions for  \( u'\) and \( v'\) are substituted into the aforementioned result. This yields the following equation:
\begin{equation}
\begin{aligned}
\mathrm{q}-\operatorname{ROFABM}^{o,h}(Q_1, Q_2, \dots, Q_m) &= \left(\frac{1}{m(m-1)} \sum_{\substack{i,j=1 \\ i \neq j}}^m Q_i^o \otimes_A Q_j^h\right)^{\frac{1}{o+h}} \\
&= \left\langle g^{-1}\left(\frac{1}{o+h} g\left(u'\right)\right), f^{-1}\left(\frac{1}{o+h} f\left(v'\right)\right)\right\rangle \\
&= \left\langle g^{-1}\left(\frac{1}{o+h} g\left(g^{-1}\left(\frac{1}{m(m-1)} \sum_{\substack{i,j=1 \\ i \neq j}}^m \left(o \cdot g(u_i) + h \cdot g(u_j)\right)\right)\right)\right), \right. \\
&\quad \left. f^{-1}\left(\frac{1}{o+h} f\left(f^{-1}\left(\frac{1}{m(m-1)} \sum_{\substack{i,j=1 \\ i \neq j}}^m \left(o \cdot f(v_i) + h \cdot f(v_j)\right)\right)\right)\right)\right\rangle
\end{aligned}
\end{equation}
\end{myproof}

In this step, we utilize the power function's scaling property to ensure that the result remains a q-rung ortho-pair fuzzy number. Through this derivation, we have demonstrated that the integrated result of the q-ROFABM operator maintains the structure of a q-rung ortho-pair fuzzy number, thereby satisfying the requirements of Theorem \ref{q-ROFABM_th1}.

\begin{theorem}\label{q-ROFWABM_th2}
 Let \( Q_{\varepsilon} = \langle u_{\varepsilon}, v_{\varepsilon} \rangle \) for \(\varepsilon = 1, 2, \dots, m\) be a set of \( Q \)-order pair fuzzy numbers. Suppose \( o, h \geq 0 \) and \( q \geq 1 \). Then the aggregated result of the q-ROFWABM operator, denoted as \( q \)-ROFWABM\(^{o,h}\left(Q_1, Q_2, \dots, Q_m\right) \), remains a \( Q \)-order pair fuzzy number. Mathematically,
\begin{equation}
\begin{aligned}
&q-\operatorname{ROFWABM}^{o, h}\left(Q_1, Q_2, \cdots, Q_m\right) \\
&= \Bigg\langle
g^{-1}\left(\frac{1}{o+h} g\left(f^{-1}\left(\frac{1}{m(m-1)}\sum_{\substack{i, j=1 \\ i \neq j}}^m f\left(g^{-1}\left(o g\left(f^{-1}\left(m w_i f\left(u_i\right)\right)\right) + h g\left(f^{-1}\left(m w_j f\left(u_j\right)\right)\right)\right)\right)\right)\right)\right), \\
& \quad f^{-1}\left(\frac{1}{o+h} f\left(g^{-1}\left(\frac{1}{m(m-1)}\sum_{\substack{i, j=1 \\ i \neq j}}^m g\left(f^{-1}\left(o f\left(g^{-1}\left(m w_i g\left(v_i\right)\right)\right) + h f\left(g^{-1}\left(m w_j g\left(v_j\right)\right)\right)\right)\right)\right)\right)\right)\Bigg\rangle
\end{aligned}
\end{equation}
\end{theorem}

\begin{myproof}
Obviously, the q-ROFABM\(^{o,h} \) operator is a special case of the q-ROFWABM\(^{o,h} \) operator. Given this relationship, the proof of this theorem is straightforward and thus omitted for brevity.
\end{myproof}

Overall, Definition \ref{q-ROFABM} provides a foundational framework for the q-ROFWABM operator, establishing the basis for weighted aggregation on q-ROFNs. Definition \ref{q-ROFWABM} extends this framework, specifying the aggregation of q-ROFNs in more complex fuzzy inference scenarios. Theorem \ref{q-ROFABM_th1} and Theorem \ref{q-ROFWABM_th2} demonstrate that the aggregation results of the q-ROFABM and q-ROFWABM operators retain the q-ROFN structure. Notably, Theorem \ref{q-ROFWABM_th2} ensures that even under complex weighted conditions, the aggregation result remains consistent with the q-ROFN form, providing theoretical support for applying the q-ROFWEBM operator in defuzzification. These definitions and theorems enable the accurate application of the q-ROFWEBM operator in multidimensional fuzzy logic inference, ensuring the reliability of the resulting driver aggressiveness assessments.

After presenting the theoretical foundations and their proofs, we will introduce the technical details of the defuzzification process. Specifically, we first define and compute the necessary intermediate variables. In particular, given the membership degrees \(\mu_{\text{BIE}_i}\) and \(\mu_{\text{BFE}_i}\) of the $i$th fuzzy number for BIE and BFE, the intermediate variables \(x_{\text{BIE}_i}\) and \(x_{\text{BFE}_i}\) can be calculated as follows:
\begin{equation}
x_{\text{BIE}_i} = \frac{2\left(\frac{\left(1+\mu_{\text{BIE}_i}^q\right)^{m w_i}-\left(1-\mu_{\text{BIE}_i}^q\right)^{m w_i}}{\left(1+\mu_{\text{BIE}_i}^q\right)^{m w_i}+\left(1-\mu_{\text{BIE}_i}^q\right)^{m w_i}}\right)^o}{\left(2-\frac{\left(1+\mu_{\text{BIE}_i}^q\right)^{m w_i}-\left(1-\mu_{\text{BIE}_i}^q\right)^{m w_i}}{\left(1+\mu_{\text{BIE}_i}^q\right)^{m w_i}+\left(1-\mu_{\text{BIE}_i}^q\right)^{m w_i}}\right)^o+\left(\frac{\left(1+\mu_{\text{BIE}_i}^q\right)^{m w_i}-\left(1-\mu_{\text{BIE}_i}^q\right)^{m w_i}}{\left(1+\mu_{\text{BIE}_i}^q\right)^{m w_i}+\left(1-\mu_{\text{BIE}_i}^q\right)^{m w_i}}\right)^o}
\end{equation}
\begin{equation}
    x_{\text{BFE}_i} = \frac{2\left(\frac{\left(1+\mu_{\text{BFE}_i}^q\right)^{m w^{'}_{i}}-\left(1-\mu_{\text{BFE}_i}^q\right)^{m w^{'}_{i}}}{\left(1+\mu_{\text{BFE}_i}^q\right)^{m w^{'}_{i}}+\left(1-\mu_{\text{BFE}_i}^q\right)^{m w^{'}_{i}}}\right)^h}{\left(2-\frac{\left(1+\mu_{\text{BFE}_i}^q\right)^{m w^{'}_{i}}-\left(1-\mu_{\text{BFE}_i}^q\right)^{m w^{'}_{i}}}{\left(1+\mu_{\text{BFE}_i}^q\right)^{m w^{'}_{i}}+\left(1-\mu_{\text{BFE}_i}^q\right)^{m w^{'}_{i}}}\right)^h+\left(\frac{\left(1+\mu_{\text{BFE}_i}^q\right)^{m w^{'}_{i}}-\left(1-\mu_{\text{BFE}_i}^q\right)^{m w^{'}_{i}}}{\left(1+\mu_{\text{BFE}_i}^q\right)^{m w^{'}_{i}}+\left(1-\mu_{\text{BFE}_i}^q\right)^{m w^{'}_{i}}}\right)^h}
\end{equation}
where \( w_i \) represents the weight of the \( i \)th fuzzy number pair for BIE, \( m \) is the total number of fuzzy number pairs, whereas \( s \) and \( h \) are the smoothness parameters of the q-ROFWEBM operator.

Similarly, the non-membership degrees \( \nu_{\text{BIE}_i} \) and \( \nu_{\text{BFE}_i} \) yield the following expressions for the intermediate variables \( y_{\text{BIE}_i} \) and \( y_{\text{BFE}_i} \). Formally,
\begin{equation}
   y_{\text{BIE}_i} = \frac{\left(1+\frac{2 \nu_{\text{BIE}_i}^{q m w_i}}{\left(2 - \nu_{\text{BIE}_i}^q\right)^{m w_i} + \nu_{\text{BIE}_i}^{q m w_i}}\right)^o - \left(1-\frac{2 \nu_{\text{BIE}_i}^{q m w_i}}{\left(2 - \nu_{\text{BIE}_i}^q\right)^{m w_i} + \nu_{\text{BIE}_i}^{q m w_i}}\right)^o}{\left(1+\frac{2 \nu_{\text{BIE}_i}^{q m w_i}}{\left(2 - \nu_{\text{BIE}_i}^q\right)^{m w_i} + \nu_{\text{BIE}_i}^{q m w_i}}\right)^o + \left(1-\frac{2 \nu_{\text{BIE}_i}^{q m w_i}}{\left(2 - \nu_{\text{BIE}_i}^q\right)^{m w_i} + \nu_{\text{BIE}_i}^{q m w_i}}\right)^o}
\end{equation}
\begin{equation}
   y_{\text{BFE}_i} = \frac{\left(1+\frac{2 \nu_{\text{BFE}_i}^{q m w^{'}_{i}}}{\left(2 - \nu_{\text{BFE}_i}^q\right)^{m w^{'}_{i}} + \nu_{\text{BFE}_i}^{q m w^{'}_{i}}}\right)^h - \left(1-\frac{2 \nu_{\text{BFE}_i}^{q m w^{'}_{i}}}{\left(2 - \nu_{\text{BFE}_i}^q\right)^{m w^{'}_{i}} + \nu_{\text{BFE}_i}^{q m w^{'}_{i}}}\right)^h}{\left(1+\frac{2 \nu_{\text{BFE}_i}^{q m w^{'}_{i}}}{\left(2 - \nu_{\text{BFE}_i}^q\right)^{m w^{'}_{i}} + \nu_{\text{BFE}_i}^{q m w^{'}_{i}}}\right)^h + \left(1-\frac{2 \nu_{\text{BFE}_i}^{q m w^{'}_{i}}}{\left(2 - \nu_{\text{BFE}_i}^q\right)^{m w^{'}_{i}} + \nu_{\text{BFE}_i}^{q m w^{'}_{i}}}\right)^h}
\end{equation}
where \( w_{i} \) and \( w^{'}_{i} \) are the weight of \( i \)-th fuzzy number pairs for BIE and BFE, respectively.

Considering real-world driving, future trajectories often depend heavily on the most recent observations, with more immediate observations typically having a greater impact on future trajectories.
To account for the time-series effects during fuzzy inference, we introduce a time decay factor that dynamically adjusts the weights of the fuzzy number pairs across different time points. The time decay factor is designed to ensure that fuzzy number pairs closer to the current moment carry more weight during defuzzification, whereas those further away are assigned less significance. The time decay factor \( \gamma_i \) for time step $\bar{t}_i$ is defined as $\gamma_i = e^{-\beta (t - \bar{t}_i)}$, where \( t \) is the current time, \( \bar{t}_i \) is the time corresponding to the $i$th fuzzy number pair. \( \beta > 0 \) is the decay rate parameter controlling how quickly the weight decreases as time \( t_i \) moves away from the current time \( t \). Consequently, the weights \(  w_i \) and \(  w_{i}^{'} \) are defined as 
$w_i= \frac{\gamma_i}{\sum_{j=1}^{m} \gamma_j} \cdot w_{\text{BIE}_i},\;\; w_{i}^{'} = \frac{\gamma_i}{\sum_{j=1}^{m} \gamma_j} \cdot w_{\text{BFE}_i}$

where \( w_{\text{BIE}_i} \) and \( w_{\text{BFE}_i} \) are hyperparameters for the $i$th fuzzy number pair of BIE and BLE, with \( w_{\text{BIE}_i} > w_{\text{BFE}_i}\) in this study. Additionally, \( \sum_{j=1}^{m} \gamma_j \) represents the sum of the decay factors across all historical observations, ensuring that the total weight is normalized to 1.

Following this, we can define the q-ROFWEBM operator for defuzzification based on the intermediate variables computed in the previous step. Formally,

\vspace{-10pt}
\begin{small}
\begin{align}
\operatorname{q\text{-}ROFWEBM}^{o, h}\left(Q_{\text{Agg}}\right) = \Bigg\langle \left(\frac{2\left(\frac{u' - u''}{u' + u''}\right)^{\frac{1}{o+h}}}{\left(2 - \frac{u' - u''}{u' + u''}\right)^{\frac{1}{o+h}} + \left(\frac{u' - u''}{u' + u''}\right)^{\frac{1}{o+h}}}\right)^{\frac{1}{q}},
\left(\frac{\left(1 + \frac{2 v'}{v' + v''}\right)^{\frac{1}{o+h}} - \left(1 - \frac{2 v'}{v' + v''}\right)^{\frac{1}{o+h}}}{\left(1 + \frac{2 v'}{v' + v''}\right)^{\frac{1}{o+h}} + \left(1 - \frac{2 v'}{v' + v''}\right)^{\frac{1}{o+h}}}\right)^{\frac{1}{q}} \Bigg\rangle 
\end{align}
\end{small}

Here, \( u' \) and \( u'' \) are intermediate variables derived from the membership degrees. Mathematically,
\begin{equation}
u^{\prime}=\left(1+\left(\frac{a_{ij}^{\prime}-a_{ij}^{\prime \prime}}{a_{ij}^{\prime}+a_{ij}^{\prime \prime}}\right)\right)^{\frac{1}{m(m-1)}}
,\;\;
u^{\prime \prime}=\left(1-\left(\frac{a_{ij}^{\prime}-a_{i j}^{\prime \prime}}{a_{i j}^{\prime}+a_{i j}^{\prime \prime}}\right)\right)^{\frac{1}{m(m-1)}}
\end{equation}
where,
\begin{small}
\begin{equation}
a' = \prod_{\substack{i,j=1 \\ i \neq j}}^m \left(1 + \frac{x_{\text{BIE}_i} x_{\text{BFE}_j}}{1 + \left(1-x_{\text{BIE}_i}\right)\left(1-x_{\text{BFE}_j}\right)}\right)^{\frac{1}{m(m-1)}}, \; \; a'' = \prod_{\substack{i,j=1 \\ i \neq j}}^m \left(1 - \frac{x_{\text{BIE}_i} x_{\text{BFE}_j}}{1 + \left(1-x_{\text{BIE}_i}\right)\left(1-x_{\text{BFE}_j}\right)}\right)^{\frac{1}{m(m-1)}}
\end{equation}
\end{small}

Similarly, \( v' \) and \( v'' \) are intermediate variables calculated based on the non-membership degrees, which can be mathematically calculated as follows:
\begin{equation}
v^{\prime}=\left(\frac{2 b_{i j}^{\prime}}{b_{i j}^{\prime}+b_{i j}^{\prime \prime}}\right)^{\frac{1}{m(m-1)}}
,
\;\; v^{\prime \prime}=\left(2-\left(\frac{2 b_{i j}^{\prime}}{b_{i j}^{\prime}+b_{i j}^{\prime \prime}}\right)\right)^{\frac{1}{m(m-1)}} 
\end{equation}
such that,
\begin{equation}
b' = \prod_{\substack{i,j=1 \\ i \neq j}}^m \left(\frac{y_{\text{BIE}_i} + y_{\text{BFE}_j}}{1 + y_{\text{BIE}_i} y_{\text{BFE}_j}}\right)^{\frac{1}{m(m-1)}}, \;\; b'' = \prod_{\substack{i,j=1 \\ i \neq j}}^m \left(2 - \frac{y_{\text{BIE}_i} + y_{\text{BFE}_j}}{1 + y_{\text{BIE}_i} y_{\text{BFE}_j}}\right)^{\frac{1}{m(m-1)}}
\end{equation}

By applying the fuzzy number pair \( Q_{\text{Agg}} \) obtained from the Max-Min aggregation to the q-ROFWEBM operator, we determine the aggressive scores of the observed traffic agents in the traffic scene, denoted as \( S = [s_0, s_1, \cdots, s_n] =\operatorname{q\text{-}ROFWEBM}^{o, h}(Q_{\text{Agg}}), \forall i\in[0,n],  s_i \in [0,1]\). A higher score indicates more aggressive behavior of the driver.

The trajectory of the ego vehicle is markedly affected by the driving behaviors of surrounding vehicles. For instance, considering a merging scene, successful lane merging is more likely when nearby drivers exhibit conservative behaviors, providing adequate space for the ego vehicle. Ignorance of the potential actions of surrounding vehicles may introduce bias into the model's estimates. Our proposed behavior-aware criteria show substantial promise in enhancing the model's environmental awareness. Specifically, high-level aggressive drivers show pronounced variations in BIE over shorter time spans, reaching notable peaks and rapid increases in BFE. In contrast, low-level aggressive drivers display more gradual changes in BIE and a slower rate of BFE increase. By utilizing the proposed fuzzy inference system, we can accurately identify and quantify these behaviors in real-time based on these behavior-aware criteria, thereby enhancing the accuracy of the trajectory prediction model.

Next, the aggressive scores $S$ from the FIS, along with the BIE $\mathcal{L}$  and BFE $\mathcal{F}$, are fed into the behavior encoder to extract frame-by-frame pyramid feature maps, which ultimately generate a high-order behavior vector $\mathbf{O}_{\text{beh}}$ for multimodal prediction of  the Multimodal Decoder.

\subsubsection{Behavior Encoder}
 This encoder is designed to provide higher-order correlations of driving behavior using the hypergraph framework \citep{yadati2019hypergcn}.
Hypergraphs, with their unique ability to connect multiple nodes through a single hyperedge, are particularly well-suited for modeling higher-order correlations. In this framework, the aggressive scores \( S \) are utilized as a criterion for establishing the hyperedge set. The node features are obtained based on BIE \( \mathcal{L} \) and BFE \( \mathcal{F} \), specifically $\hat{v}_i = \phi_{\textit{MLP}}(\mathcal{L}_i,\mathcal{F}_i)$. We define a behavior hypergraph \( \mathcal{G} = (\mathcal{V}, \mathcal{E}) \), where each hyperedge delineates a group of drivers exhibiting similar behaviors. The node set \( \mathcal{V} = \{\hat{v}_0, \hat{v}_1, \ldots, \hat{v}_n\} \) represents the traffic agents, with each node corresponding to a distinct vehicle, and \( n \) denoting the total number of surrounding vehicles. The hyperedge set \( \mathcal{E} = \{\hat{e}_L, \hat{e}_M, \hat{e}_H\} \) includes hyperedges \( \hat{e}_L \), \( \hat{e}_M \), and \( \hat{e}_H \), which connect vehicles that demonstrate aggressive behaviors at low, medium and high levels, respectively.
The hyperedge set $\mathcal{E} \in \mathbb{R}^{(n+1) \times 3}$ is the foundation for hypergraph learning and could be formulated as a matrix. We take the $e_L$ as an illustrated example. Formally,
\begin{equation}
  \hat{e}_{L,i} = 
\begin{cases}
1, & \text{if} \quad s_i < 0.3, \\
0, & \text{else}.
\end{cases}  
\end{equation}
where $\hat{e}_{L, i} =1$ represents the driver of vehicle $i$ exhibiting Low-level aggressive behaviors, and vice versa. $s_i$ denotes the score of vehicle $i$. For a given node $\hat{v} \in \mathcal{V}$, its degree is defined by $d_{hyp}(\hat{v}) = \sum_{\hat{e} \in \mathcal{E}} \omega_{hyp}(\hat{e}) h_{hyp}(\hat{v}, \hat{e})$, where $\omega_{hyp}(\hat{e})$ represents the weight of the edge $\hat{e}$. 
Similarly, for an edge $\hat{e} \in \mathcal{E}$, its degree is calculated as $\delta_{hyp}(\hat{e}) = \sum_{\hat{v} \in \mathcal{V}} h(\hat{v}, \hat{e})$. Moreover, let $\mathbf{D}_{\hat{e}}$, and 
$\mathbf{D}_{\hat{v}}$ be the diagonal matrices corresponding to the degrees of edges and vertices, respectively.

To perform convolution on a signal \( \mathbf{x} \) over the hypergraph, we first apply a Fourier transform represented as \(\Phi \). A filter \( \hat{\mathbf{g}}(\Lambda) \) is then applied to these frequency domain features. The output of the hypergraph convolution can be mathematically expressed as follows:
\begin{equation}
\hat{\mathbf{g}} \ast \mathbf{x} = \Phi \left( (\Phi^\top \hat{\mathbf{g}}) \odot (\Phi^\top \mathbf{x}) \right) = \Phi \hat{\mathbf{g}}  (\Lambda) \Phi^\top \mathbf{x},
\end{equation}
Here, \( \odot \) denotes element-wise multiplication, and \( \hat{\mathbf{g}} (\Lambda) \) can be interpreted as a spectral filter that adjusts the response to different frequency components, akin to traditional graph convolution filters. Direct computation of the Fourier transform is computationally expensive. Therefore, we utilize Chebyshev polynomials \( T_k(x) \) to approximate \( \hat{\mathbf{g}} (\Lambda) \), which avoids explicit eigen decomposition by relying on the recursive definition of polynomials. Let \(\Delta\) donates the hypergraph Laplacian matrix, \(\Lambda = \text{diag}(\lambda_1, \ldots, \lambda_n)\) is the diagonal matrix containing the corresponding non-negative eigenvalues. The \(\lambda_{\max}\) represents the largest corresponding non-negative eigenvalue. We define \( \tilde{\Delta} = \frac{2}{\lambda_{\max}} \Delta - I \). Then, the filter \( \hat{\mathbf{g}}(\Lambda) \) can be approximated as follows:
\begin{equation}
\hat{\mathbf{g}} \ast \mathbf{x} \approx \sum_{k=0}^K \theta_k T_k(\tilde{\Delta}) \mathbf{x}  \approx \frac{1}{2}  \theta_k \mathbf{D}_v^{-1/2} \mathbf{H} (\mathbf{W} + \mathbf{I}) \mathbf{D}_e^{-1} \mathbf{H}^\top \mathbf{D}_v^{-1/2} \mathbf{x}
\approx \theta_k\mathbf{D}_v^{-1/2} \mathbf{H} \mathbf{W} \mathbf{D}_e^{-1} \mathbf{H}^\top \mathbf{D}_v^{-1/2} \mathbf{x}
\end{equation}
where \( \theta_k \) are the learnable parameters and \( K \) denotes the order of the polynomial. The \( \mathbf{W} + \mathbf{I} \) represents the weights of the hyperedges. Here, \( \mathbf{W} \) is initialized as an identity matrix, indicating equal weight for all hyperedges initially. The convolution operation can be further simplified using the node and hyperedge degree matrices. We can then define:
\begin{equation}
    \mathbf{\bar{X}} = \left[\{\mathcal{I}^{t-t_h}, \mathcal{F}^{t-t_h}\}, \{\mathcal{I}^{t-t_h+1}, \mathcal{F}^{t-t_h+1}\}, \ldots, \{\mathcal{I}^{t}, \mathcal{F}^{t}\}\right]\
\end{equation}
\begin{equation}
    \mathbf{O}_{\text{beh}} = \mathbf{D}_v^{-1/2} \mathbf{H} \mathbf{W}_{\text{hyp}} \mathbf{D}_e^{-1} \mathbf{H}^\top \mathbf{D}_v^{-1/2} \mathbf{\bar{X}} \Theta
\end{equation}
where \(\bm{D}_{e}\) and \(\bm{D}_{v}\) are the diagonal matrices of edge and vertex degrees, respectively. \( \Theta \) is a learnable parameter, \( \mathbf{W}_{\text{hyp}}\) is the diagonal hyperedge weight matrix. 

In a nutshell, the workflow of the Behavior-aware Module operates as follows: Given the historical states of the ego vehicle and its observed surrounding agents, we first construct Dynamic Geometric Graphs to model the dynamic interactions among traffic agents. Next, centrality measures—dynamic degree centrality, dynamic closeness centrality, and eigenvector centrality—are computed for each agent at every time step to quantify their interactions and influence. Based on these centrality measures, we derive two behavior-aware criteria:  BIE \(\mathcal{I}\) and BFE \(\mathcal{F}\). These criteria, calculated over the entire historical observation window, provide a comprehensive view of interaction dynamics. The criteria are then aggregated into a driving aggressiveness score \(S\) using the Fuzzy Inference System, through the processes of fuzzification, fuzzy inference, and defuzzification.  Finally,
the aggressiveness score \(S\), along with the BIE and BFE, are fed into the behavior encoder. This encoder adopts a hierarchical hypergraph-based method to generate the high-order behavior vector \(\mathbf{O}_{\text{beh}}\), effectively capturing complex behavioral patterns. The behavior vector \(\mathbf{O}_{\text{beh}}\) is then passed to the Multimodal Decoder, where it serves as a critical input for producing context-aware trajectory predictions.

\subsection{Interaction-aware Module}\label{Interaction-aware Pooling Module_0}
The influence of surrounding vehicles is a critical factor in the decision-making process of the ego vehicle. To align with the human-like understanding of this mutual influence, we implemented a pooling mechanism based on a polar coordinate system. This approach addresses the limitations of the traditional Cartesian coordinate system \citep{wang2023wsip,liao2024human}, which often struggles to effectively model unstructured road topologies. Interaction is captured through the interplay of both spatial and temporal dimensions.  To address this, we introduce the Interaction-aware Module, which consists of a Gated Recurrent Unit (GRU) encoder, Positional Encoding, and an Interaction Encoder. The GRU encoder is responsible for embedding the input historical observations $\bm{X}$ into a sequence of tokens. Positional encoding is then applied to generate position vectors $\mathbf{O}_{\text{pos}}$. Finally, we propose a novel attention mechanism in the interaction encoder to produce the high-level interaction vectors $\mathbf{O}_{\text{int}}$.  For the sake of brevity, additional details of this module are provided in \textbf{Appendix \ref{Interaction-aware Module}}.

\subsection{Multimodal Decoder}\label{Decoder Moudle} 
The decoder, based on a multimodal GMM, is responsible for integrating the behavior and interaction vectors to generate multimodal future trajectories. Specifically, the behavior $\mathbf{O}_{\text{beh}}$ and interaction $\mathbf{O}_{\text{int}}$ vectors are combined through a fully connected layer and group normalization (GN) computation. This composite vector is then fed into a GRU to produce a high-dimensional fusion feature. Subsequently, this feature is passed through a softmax activation function, followed by a fully connected layer, which generates a probability distribution over the potential future trajectories of the ego vehicle. Formally, $\bm{Y}_{0}^{t:t+t_{f}} = \phi_{\text{MLP}}\left(\phi_{\text{softmax}}\\\phi_{\text{MLP}}(\left(\phi_{\text{GRU}}(\phi_{\text{GN}}(( \phi_{\text{MLP}}(\mathbf{O}_{\text{beh}})\odot\phi_{\text{MLP}}(\mathbf{O}_{\text{int}})))))\right)\right)$
where is the GN computation. In particular, the fully connected layers within the decoder are designed with independent parameter sets, ensuring that behavior and interaction vectors are learned distinctly. This separation enhances the model’s ability to capture the unique contributions of each modality, preventing information entanglement that could degrade prediction quality. Overall, this multimodal architecture not only generates trajectory predictions for different maneuvers but also quantifies their confidence levels, facilitating more reliable decision-making under uncertainty.

In particular, the fully connected layers within the decoder do not share parameters.  These distinct parameter sets allow the model to better capture the unique contributions of behavior and interaction vectors. Overall, this multimodal structure not only provides predictions for different maneuvers but also quantifies their confidence levels, thereby supporting decision-making in the face of uncertainty.

\section{Experiments}\label{Experiments}
In this section, we demonstrate the effectiveness of the HiT framework from four key perspectives. Firstly, the overall evaluation results showcase our model's accuracy and efficiency across various complex scenarios, especially under conditions with limited training data. Secondly, ablation studies are conducted to evaluate the contribution of each component in HiT, underscoring their individual impacts. Thirdly, through detailed case studies, we empirically validate the effectiveness of the proposed behavior-aware approach, which is in line with the theoretical foundations previously discussed. Lastly, we offer a series of qualitative results in various kinds of traffic conditions to further illustrate HiT's inference process in real-world driving scenarios.

\subsection{Experiment Setups}
\subsubsection{Datasets}
We evaluated the efficacy of HiT using five datasets: NGSIM, HighD, RounD, ApolloScape, and MoCAD++. These datasets, drawn from diverse real-world scenes—including highways, roundabouts, and urban roads—provide a comprehensive testing ground for our HiT model.

\subsubsection{Data Segmentation}
These datasets were partitioned into training, validation, and test sets using standard sampling. We refer to the complete test set as the \textit{overall} test set. The trajectories for the NGSIM, HighD, and MoCAD++ datasets were divided into 8-second intervals. The first 3 seconds served as the trajectory history ($t_h=3$) for input, and the following 5 seconds represented the ground truth ($t_f = 5$) for output. For the RounD dataset, the trajectories were divided into 6-second chunks with $t_h=2$ and $t_f = 4$. In addition, for the ApolloScape dataset, we adhere to the protocol set forth by the ApolloScape Trajectory Forecasting Challenge. Specifically, we predict 3-second future trajectories (\(t_f = 3\)) for a diverse range of traffic agents, including vehicles, pedestrians, and cyclists, using the initial 3-second observation period (\(t_h = 3\)) as the basis for these predictions. To delve deeper into our model's performance, the NGISM dataset was further split based on distinct vehicular maneuvers, including no lane-change (\textit{keep}), on-ramp lane merging  (\textit{merge}), right lane-change (\textit{right}), and left lane-change (\textit{left}). This subset, termed the \textit{maneuver-based} test set, allows for a more granular examination of HiT's capabilities across different actions. In addition, we conduct the experiments based on a small model with 64 hidden units, termed HiT (S).

\subsubsection{Evaluation Metrics}
To ensure a fair comparison with other baselines, we adhere to the metric settings specified in benchmark datasets. For NGSIM, HighD, RounD, and MoCAD++, we employ the Root Mean Square Error (RMSE) as the evaluation metric. For the ApolloScape, Average Displacement Error (ADE) and Final Displacement Error (FDE) are utilized as metrics.

\subsubsection{Loss Function}\label{Loss function}
To train our model, we employ two distinct loss functions: the maneuver loss \(\mathcal{L}_\textit{man}\) and the trajectory loss \(\mathcal{L}_\textit{tra}\). The maneuver loss \(\mathcal{L}_\textit{man}\) guides the model in accurately predicting the intended maneuver, whereas the trajectory loss \(\mathcal{L}_\textit{tra}\) ensures that the predicted trajectory closely aligns with the ground truth, thereby enhancing the overall predictive accuracy. For the NGSIM, HighD, RounD, and MoCAD++ datasets, we treat the output trajectory as a bivariate Gaussian distribution. Given the ground-truth position coordinates \(\hat{p}_{0}^{t_k}= (\hat{\rho}_{0}^{t_k}, \hat{\theta}_{0}^{t_k})\), and the final predicted trajectory output by HiT, denoted as \({p}_{0}^{t_k }=\{\rho_{0}^{t_k},\theta_{0}^{t_k}\}\), the maneuver loss $\mathcal{L}_\textit{man}$ uniformly across all datasets can be formally calculated as follows:

\vspace{-10pt}
{\begin{footnotesize}
\begin{equation}\label{nll}
\begin{aligned}
\mathcal{L}_{\textit{man}} = \sum_{t_k=t+1}^{t+t_f} { \log \left(2\pi \sigma_{t_k, \rho} \sigma_{t_k, \theta} \sqrt{1-\eta_{t_k}^2}\right)} +
\left\{\frac{1}{2(1-\eta_{t_k}^2)}\left[\left(\frac{\hat{\rho}_{0}^{t_k}-{\rho}_{0}^{t_k}}{\sigma_{t_k, \rho}}\right)^2\right. \right. 
-2\eta_{t_k} \frac{\hat{\rho}_{0}^{t_k}-\rho_{0}^{t_k}}{\sigma_{t_k, \rho}}\frac{\hat{\theta}_{0}^{t_k}-\theta_{0}^{t_k}}{\sigma_{t_k, \theta}}
\left.\left.+\left(\frac{\hat{\theta}_{0}^{t_k}-\theta_{0}^{t_k}}{\sigma_{t_k, \theta}}\right)^2 \right]\right \}
\end{aligned}
\end{equation}
\end{footnotesize}}
where \(\sigma_{t_k, \rho}\), \(\sigma_{t_k, \theta}\) are the means and variances of \(\hat{\rho}_{0}^{t_k}\) and \({\rho}_{0}^{t_k}\), and \(\eta_{t_k}\) is the correlation coefficient between \(\hat{\rho}_{0}^{t_k}\) and \({\rho}_{0}^{t_k}\).
This metric is particularly valuable for assessing the fidelity of trajectory predictions relative to expected maneuvers, ensuring the reliability of trajectory forecasting within action-based models. 
Correspondingly, we employ the MSE metric as the trajectory loss \(\mathcal{L}_\textit{tra}\), which quantifies the average Euclidean distance between the predicted and ground truth trajectories, serving as a comprehensive measure of predictive accuracy. Notably, HiT is trained using only the trajectory loss \(\mathcal{L}_\textit{t}\) for ApolloScape, which is specifically designed for multi-agent trajectory prediction in a single forward pass. To ensure consistency with evaluation protocols, we adopted a tailored loss function consistent with existing studies for fair comparison. We leave more details in \textbf{Appendix \ref{Loss}}.

\begin{table}[t]
  \centering
\caption{Comparison of HiT with baseline models. ``Pos.'', ``Spe.'', and  `Acc.'' denote the 2D positional coordinates, speed, and acceleration of agents, respectively. ``VAE'' denotes the Variational Auto Encoder, ``GAN'' is the Generative Adversarial Network, ``MHA'' is the Multi-head Attention mechanism, ``CNN'' is the Convolutional Neural Network, and ``HGCN'' IS the Hypergraph Convolutional Network. Correspondingly, ``TCN'' is the Temporal Convolution Network.  Correspondingly, the ``Loss'' column specifies the loss functions used during training, including mean squared error (MSE), negative log-likelihood (NLL), and L2 Loss, Maximum Likelihood Estimation (MLE), and kullback-leibler divergence (KL) loss. The target agents for prediction include ``Veh.'' for various types of vehicles (cars, trucks, trailers, vans, buses), ``Ped.'' for pedestrians, and ``Bic.'' for bicyclists.}
     \setlength{\tabcolsep}{3.5mm}
   \resizebox{\linewidth}{!}{
    \begin{tabular}{cccccccc}
    \toprule
    Model & Input & Encoder & Interaction Module & Decoder & Loss  & Targe Agent & Multimodal \\
    \midrule
    M-LSTM \citep{deo2018multi} & Pos.   & LSTM  & MLP   & LSTM  & MSE, NLL & Veh., Ped. & Yes \\
    S-GAN \citep{gupta2018social} & Pos.   & LSTM  & MLP   & LSTM  & MSE   & Veh., Ped. & No \\
    CS-LSTM \citep{deo2018convolutional} & Pos., Spe., Acc. & LSTM  & MLP   & LSTM, MLP & MSE, NLL & Veh.  & Yes \\
    NLS-LSTM \citep{messaoud2019non} & Pos., Spe., Acc. & LSTM  & CNN   & LSTM  & MSE   & Veh.  & No \\
    MFP \citep{tang2019multiple} & Pos., Spe., Acc. & RNN   & RNN   & RNN   & MLE, KL & Veh.  & No \\
    DN-IRL \citep{fernando2019neighbourhood} & Pos., Spe., Acc. & LSTM  & CNN   & CNN   & MSE, NLL & Veh., Ped. & Yes \\
    CF-LSTM \citep{xie2021congestion} & Pos., Spe., Acc. & GCN   & VAE   & LSTM  & MSE, NLL & Veh.  & Yes \\
    STDAN \citep{chen2022intention} & Pos., Spe., Acc. & LSTM, MLP & Attention & Attention, MLP & NLL   & Veh.  & Yes \\
    DRBP\citep{gao2023dual}  & Pos., Spe., Acc. & Transformer & Transformer & LSTM, MLP & MSE   & Veh.  & No \\
    WSiP \citep{wang2023wsip} & Pos., Spe., Acc. & LSTM, MLP & CNN, MLP & LSTM  & MSE, NLL & Veh.  & Yes \\
    GaVa \citep{liao2024human} & Pos., Spe., Acc. & GRU, CNN & GAT   & Transformer & MSE, NLL & Veh.  & Yes \\
    HLTP++ \citep{liao2024less} & Pos., Spe., Acc. & LSTM  & FA-SNN, Transformer & MHA,MLP & MSE, NLL & Veh.  & Yes \\
    TrafficPredict \citep{ma2019trafficpredict} & Pos., Spe., Acc. & LSTM  & LSTM  & LSTM  & MSE, NLL & Veh., Bic., Ped. & Yes \\
    GRIP++\citep{li2020gripplus} & Pos., Spe., Acc. & CNN   & GCN   & GRU   & MSE   & Veh., Bic., Ped. & No \\
    TPNet \citep{fang2021tpnet} & Pos., Spe., Acc. & CNN   & MLP   & LSTM, MLP & MSE   & Veh.  & Yes \\
    S2TNet \citep{chen2022s2tnet} & Pos., Spe., Acc. & Transformer & Transformer, CNN & Transformer & L2 Loss & Veh., Bic., Ped. & Yes \\
    AI-TP \citep{zhang2022ai} & Pos., Spe., Acc. & CNN, GAT & GAT   & Transformer & MSE, L2 Loss   & Veh., Bic., Ped. & No \\
    MSTG \citep{mstg}  & Pos., Spe., Acc. & LSTM  & GCN, TCN & LSTM  & MSE   & Veh., Bic., Ped. & No \\
    TP-EGT \citep{tp-egt} & Pos., Spe., Acc. & LSTM  & Transformer & LSTM, MLP & MSE, L2 Loss   & Veh., Bic., Ped. & Yes \\
    \midrule
    \textbf{HiT} & Pos., Spe., Acc. & HGCN, GRU & MHA & GRU, MLP & MSE, NLL & Veh., Bic., Ped. & Yes \\
    \bottomrule
    \end{tabular}%
    }
  \label{tab:baseline}%
\end{table}%

\begin{table}[htbp]
  \centering
     \caption{Evaluation results for HiT and baseline models on the \textit{overall} test set across different prediction horizons. Note: RMSE (m) is used as the evaluation metric, with lower values indicating better performance; ‘-’ indicates values not provided. \textbf{Bold} values denote the best performance, and \underline{underlined} values denote the second-best performance in each category. AVG represents the average RMSE across all horizons.}\label{Table1}
     \setlength{\tabcolsep}{8mm}
   \resizebox{0.98\linewidth}{!}{
    \begin{tabular}{c|ccccccc}
    \toprule
    \multicolumn{1}{c}{\multirow{2}[2]{*}{Dataset}} & \multirow{2}[3]{*}{Model} & \multicolumn{6}{c}{Prediction Horizon (s)} \\
\cmidrule{3-8}    \multicolumn{1}{c}{} &       & 1     & 2     & 3     & 4     & 5 & AVG\\
    \midrule
    \multirow{12}[21]{*}{NGSIM} 
          & M-LSTM \citep{deo2018multi} & 0.58  & 1.26  & 2.12  & 3.24  & 4.66 & 2.37 \\
          & S-GAN \citep{gupta2018social} & 0.57  & 1.32  & 2.22  & 3.26  & 4.40 & 2.35 \\
          & CS-LSTM \citep{deo2018convolutional} & 0.61  & 1.27  & 2.09  & 3.10  & 4.37 & 2.29 \\
          & NLS-LSTM \citep{messaoud2019non} & 0.56  & 1.22  & 2.02  & 3.03  & 4.30 & 2.23 \\
          & MFP \citep{tang2019multiple} & 0.54  & 1.16  & 1.89  & 2.75  & 3.78 & 2.02 \\
          & DN-IRL \citep{fernando2019neighbourhood} & 0.54  & 1.02  & 1.91  & 2.43  & 3.76 & 1.93 \\
          & MHA-LSTM \citep{messaoud2021attention} & 0.41  & 1.01  & 1.74  & 2.67  & 3.83 & 1.91 \\
          & CF-LSTM \citep{xie2021congestion} & 0.55  & 1.10  & 1.78  & 2.73  & 3.82  & 1.99 \\

          & STDAN \citep{chen2022intention} & \underline{0.39}  & 0.96  & 1.61  & 2.56 & 3.67 & 1.84 \\
          & DRBP\citep{gao2023dual} & 1.18  & 2.83  & 4.22  & 5.82  & -  & 3.51 \\
          & WSiP \citep{wang2023wsip} & 0.56  & 1.23  & 2.05  & 3.08  & 4.34 & 2.25 \\
          & GaVa \citep{liao2024human} & 0.40 & 0.94 & 1.52 & 2.24 & 3.13  & 1.65\\
          & HLTP++ \citep{liao2024less} & 0.46 & 0.98 & 1.52 & 2.17 & 3.02  & \underline{1.63} \\

          & \textbf{HiT (25\%)} & 0.46 & 0.99  & 1.54  & 2.18  & 3.04 & 2.01 \\
         & \textbf{HiT (S)} & {0.42}  & \underline{0.93}  & \underline{1.46}  & \underline{2.14}  & \underline{2.95}  & {1.59}\\

          & \textbf{HiT} & \textbf{0.38} & \textbf{ 0.90 } & \textbf{  1.42 } & \textbf{ 2.08 } & \textbf{ 2.87 }  & \textbf{1.53}\\

    \midrule
    \multirow{7}[37]{*}{HighD} 

    &S-GAN \citep{gupta2018social}& 0.30  & 0.78  & 1.46  & 2.34  & 3.41  &1.69 \\
    &CS-LSTM \citep{deo2018convolutional}& 0.22  & 0.61  & 1.24  & 2.10  & 3.27 &1.48 \\

    &NLS-LSTM \citep{messaoud2019non}& 0.20  & 0.57  & 1.14  & 1.90  & 2.91 &1.34\\

    &MHA-LSTM \citep{messaoud2021attention}& 0.19  & 0.55  & 1.10  & 1.84  & 2.78 &1.29 \\
    &CF-LSTM \citep{xie2021congestion}& 0.18  & 0.42  & 1.07  & 1.72  & 2.44 & 1.17  \\
    &EA-Net \citep{cai2021environment} & 0.15  & 0.26  & 0.43  & 0.78  & 1.32  &0.59 \\

    &iNATran \citep{chen2022vehicle}& \textbf{0.04}  & \textbf{0.05}  & \textbf{0.21}  & 0.54  & 1.10 &\underline{0.39}\\
    &STDAN \citep{chen2022intention}& 0.19  & 0.27  & 0.48  & 0.91  & 1.66  &0.70 \\
    
    &WSiP \citep{wang2023wsip}& 0.20  & 0.60  & 1.21  & 2.07  & 3.14 &1.44 \\
    &DRBP\citep{gao2023dual}& 0.41  & 0.79  & 1.11  & 1.40  & -  & 0.92\\
    
    & GaVa \citep{liao2024human} & 0.17 & 0.24 & 0.42 & 0.86 & 1.31  & 0.93 \\
    & Wave \citep{liao2024physics} & 0.14 & 0.24 & 0.37 & \underline{0.50} & \underline{0.72} & \underline{0.39} \\
    
    & \textbf{HiT (25\%)} & 0.19  & 0.37  & 0.68  & 0.89  & 1.33 &0.83  \\
    & \textbf{HiT (S)} & \underline{0.07}  & 0.27  & 0.34 & 0.51  & 0.94 &0.42 \\
    & \textbf{HiT} & 0.08  & \underline{0.13}  & \underline{ 0.22 } & \textbf{ 0.39 } & \textbf{ 0.61 }  &\textbf{0.28}\\
     \midrule
    \multirow{7}[22]{*}{RounD}     
    &S-LSTM \citep{alahi2016social} & 0.94 & 1.82 & 3.43 & 5.21& - & 2.85 \\
    &S-GAN \citep{gupta2018social} & 0.72 & 1.57 & 3.01 & 4.78& - & 2.52 \\

    &CS-LSTM \citep{deo2018convolutional} & 0.71 & 1.21 & 2.09 & 3.92& - & 1.98 \\
    &NLS-LSTM \citep{messaoud2019non} & 0.62 & 0.96 & 1.91 & 3.48& - & 1.74 \\
    &MATH \citep{hasan2021maneuver} & 0.38 & 0.80 & 1.76 & 3.08& - & 1.51 \\
    &MHA-LSTM \citep{messaoud2021attention} & 0.62 & 0.98 & 1.88 & 3.65& - & 1.78 \\

    &CF-LSTM \citep{xie2021congestion} & 0.51 & 0.87 & 1.79 & 3.14& - & 1.57 \\
    &STDAN \citep{chen2022intention} & 0.35 & 0.77 & 1.74 & 2.92& - & 1.46 \\
    &WSiP \citep{wang2023wsip} & 0.52 & 0.99 & 1.88 & 3.07& - & 1.61 \\

    & \textbf{HiT (25\%)} & 0.44 & 0.82 & 2.10 & 3.22 & - & 1.64 \\
     & \textbf{HiT (S)} & \underline{0.33}  & \underline{0.71}  & \underline{1.60}  & \underline{2.75}  & - & \underline{1.34} \\
    & \textbf{HiT} & \textbf{0.21}  & \textbf{0.52}  & \textbf{1.40}  & \textbf{2.41}  & - & \textbf{1.13} \\
     \midrule
    \multirow{7}[20]{*}{MoCAD++}

    &S-GAN \citep{gupta2018social} & 1.82  & 2.38  & 3.31  & 3.96  & 4.75 & 3.24 \\
    &CS-LSTM \citep{deo2018convolutional} & 1.47  & 2.05  & 2.97  & 3.72  & 4.53 & 2.95 \\
    &NLS-LSTM \citep{messaoud2019non} & 0.99  & 1.36  & 2.13  & 2.93  & 4.12 &2.31 \\
    &MHA-LSTM \citep{messaoud2021attention} & 1.36  & 1.48  & 2.65  & 3.48  & 4.38  & 2.67\\
    &CF-LSTM \citep{xie2021congestion} & 0.81  & 1.11  & 1.81  & 2.63  & 3.58 & 1.99 \\
    &STDAN \citep{chen2022intention} & 0.86  & 1.01  & 1.78  & 2.73  & 3.61 & 2.00  \\
    &WSiP \citep{wang2023wsip} & 0.67  & 0.90  & 1.69  & 2.53  & 3.36 &1.83  \\
    & HLTP++ \citep{liao2024less} & 0.68 & 0.92 & 1.69 & 2.64 & 3.50  & 1.89 \\
    &GaVa \citep{liao2024human} & \underline{0.59} & \underline{0.82} & \underline{1.46} & \underline{2.26} & \underline{3.03} & \underline{1.63}\\

     & \textbf{HiT (25\%)}& 0.77  & 0.98  & 1.92  & 2.78  & 3.52 &1.99 \\
     & \textbf{HiT (S)} & 0.63  & 0.88 & 1.52 & 2.49  & 3.30  & 1.76 \\
    & \textbf{HiT} & \textbf{0.32}  & \textbf{0.75}  & \textbf{1.36}  & \textbf{2.18}  & \textbf{2.82}  & \textbf{1.48}\\
    \bottomrule
    \end{tabular}%
  \label{tab:addlabel}%
  }
\end{table}%

\subsubsection{Training and Implementation Details}\label{Training} 
Our model was trained on an NVIDIA A40 GPU with 48GB of memory. The training process employed a batch size of 128, with the model trained for 60 epochs on the ApolloScape dataset and 12 epochs on the other datasets. The training time for HiT varies by dataset: for the NGSIM and HighD datasets, each epoch takes approximately 45 minutes. For the MoCAD++ dataset, which involves more traffic agents and diverse data, training time increases to nearly 2 hours per epoch. On the ApolloScape dataset, each epoch takes approximately 2–3 minutes. Moreover, we adopted a dynamic learning rate strategy, initially set at $10^{-3}$ and gradually reducing to $10^{-5}$. The Adam optimizer was employed, coupled with the Cosine Annealing Warm Restarts (CAWR) scheduler to manage the learning rate adjustments. We provide a detailed, modular description of the model parameters in \textbf{Supplementary Information C}.

\subsection{Experimental Results} 
Table~\ref{tab:baseline} presents a comprehensive comparison of our proposed HiT with baseline methods across key components, including input features, encoder type, interaction modules, decoder architecture, and training loss functions. HiT is unique in that it introduces several novel design elements. The behavior-aware module dynamically models driving behaviors using Behavior-aware Criteria, providing a more context-aware and interpretable representation. Additionally, HiT’s hypergraph-based behavior encoder captures higher-order relationships among diverse traffic agents, surpassing baseline models that are constrained to pairwise interactions. The hybrid architecture of HiT, combining HGCN, GRU, and attention mechanisms, enables efficient multi-agent trajectory prediction in a single forward pass, whereas its multitask loss function balances multimodal maneuver and trajectory prediction for improved accuracy and robustness.

\textbf{Performance Evaluation on the \textit{Overall} Test Set.} To validate the robustness of our model across various scenarios, we conducted experiments on the NSGIM, HighD, RounD, MoCAD++, and ApolloScape datasets. The results were compared against both SOTA models from the past five years and several classical models. The detailed comparisons are presented in Tables \ref{tab:addlabel} and \ref{tab:apollo}. We provide a comprehensive analysis of the results for each dataset in \textbf{Appendix \ref{Overall Experimental Results}}.

\begin{table*}[t]
\centering
\setlength{\tabcolsep}{2mm}
\caption{{Evaluation results on the ApolloScape dataset. $\textit{ADE}_{v/p/b}$ and $\textit{FDE}_{v/p/b}$ represent the ADE and FDE metrics for vehicles, pedestrians, and bicycles, respectively.}}
  \resizebox{0.9\linewidth}{!}{
\begin{tabular}{c|c|ccc|c|ccc}
\bottomrule
Model  & WSADE & ADEv & ADEp & ADEb & WSFDE & FDEv & FDEp & FDEb \\
\hline
TrafficPredict \citep{ma2019trafficpredict}  & 8.5881 & 7.9467 & 7.1811 & 12.8805 & 24.2262 & 12.7757 & 11.1210 & 22.7912 \\
GRIP++\citep{li2020gripplus} & 1.2588 & 2.2400 & 0.7142 & 1.8024 & 2.3631 & 4.0762 & 1.3732 & 3.4155 \\
TPNet \citep{fang2021tpnet}  & 1.2800 & 2.2100 & 0.7400 & 1.8500 & 2.3400 & 3.8600 & 1.4100 & 3.4000 \\
S2TNet \citep{chen2022s2tnet}  & 1.1679 & {1.9874} & {0.6834} & 1.7000 & 2.1798 & 3.5783 & 1.3048 & 3.2151 \\
AI-TP \citep{zhang2022ai} & \underline{1.1544} & 1.9878 & \underline{0.6684} & 1.6780 & {2.1297} & \underline{3.5246} & {1.2667} & 3.1370 \\
MSTG \citep{mstg}   & 1.1546 & \underline{1.9850} & {0.6710} & \underline{1.6745} & \underline{2.1281} & 3.5842 & \underline{1.2652} & \underline{3.0792} \\
TP-EGT \citep{tp-egt} & 1.1900 & 2.0500 & 0.7000 & 1.7200 & 2.1400 & {3.5300} & {1.2800} & 3.1600 \\
\hline
\textbf{HiT} & \textbf{1.1246} & \textbf{1.8023} & \textbf{0.6442} & \textbf{1.5211} & \textbf{1.9134} & \textbf{3.1871} & \textbf{1.2062} & \textbf{2.8113} \\
\toprule
\end{tabular}
}
\label{tab:apollo}
\end{table*}

\textbf{Performance Evaluation on the \textit{Maneuver-based} Test Set.} To gain a comprehensive understanding of our proposed model's performance, we conducted additional evaluations on the \textit{maneuver-based} test set. The detailed evaluation results are provided in \textbf{Appendix \ref{Maneuver}}.

\textbf{Comparative Analysis of Model Performance and Model Efficiency.}
We benchmarked the performance and complexity of our HiT and HiT (S) models against other models, as presented in Table \ref{weight}. Both HiT and HiT (S) demonstrate superior performance across all metrics while maintaining significantly lower complexity. See \textbf{Appendix \ref{Model Efficiency}} for analyses of these evaluation results.

\textbf{Comparative Analysis of Real-time Performance.}
In real-world autonomous driving, real-time inference performance is crucial, as any delay in decision-making can have significant safety implications. Consequently, assessing the efficiency of trajectory prediction models is vital to ensuring their practical applicability in real-world systems. We evaluated the real-time performance of HiT, and the results and analyses are presented in \textbf{Appendix \ref{Real-time Performance}}

\begin{table}[tb]
  \centering
  \caption{Performance comparison across varying levels of driving aggression and prediction horizons. Results are presented for Low, Medium, and High aggression subsets at different time horizons (1-5 seconds). The table contrasts the HiT model and its variant without the behavior-aware module (HiT (-BA)) against baseline models, with lower values indicating better performance in each category.}
    \setlength{\tabcolsep}{2mm}
   \resizebox{0.85\linewidth}{!}{
    \begin{tabular}{cccccccccccccccc}
    \toprule
    Dataset & \multicolumn{5}{c}{Low}       & \multicolumn{5}{c}{Medium}    & \multicolumn{5}{c}{High} \\
        \cmidrule(r){1-1} \cmidrule(r){2-6}  \cmidrule(r){7-11}  \cmidrule(r){12-16} 
   \multirow{1}{*}{Model} & \multicolumn{5}{c}{Horizon (s)}       & \multicolumn{5}{c}{Horizon (s)}       & \multicolumn{5}{c}{Horizon (s)} \\
    \cmidrule(r){1-1}
\cmidrule(r){2-6} \cmidrule(r){7-11}  \cmidrule(r){12-16}   & 1     & 2     & 3     & 4     & 5     & 1     & 2     & 3     & 4     & 5     & 1     & 2     & 3     & 4     & 5 \\
    WSiP \citep{wang2023wsip} & 0.61  & 1.25  & 2.15  & 3.19  & 4.41  & 0.45  & 1.20  & 1.96  & 2.97  & 4.27  & 0.90  & 1.31  & 2.20  & 3.30  & 4.47  \\
    STDAN \citep{chen2022intention}& 0.42  & 0.99  & 1.62  & 2.61  & 3.69  & 0.32  & 0.89  & 1.52  & 2.49  & 3.59  & 0.58  & 1.15  & 1.95  & 2.72  & 3.95  \\
    GaVa \citep{liao2024human} & 0.46  & 0.97  & 2.47  & 3.16  & 3.30  & 0.34  & 0.54  & 0.78  & 1.52  & 2.93  & 0.50  & 2.40  & 2.48  & 3.19  & 3.55  \\
    HLTP++ \citep{liao2024less}& 0.46  & 1.05  & 1.56  & 2.16  & 3.21  & 0.33  & 0.90  & 1.45  & 2.05  & 2.85  & 0.91  & 1.22  & 1.68  & 2.61  & 3.31  \\
    \cmidrule(r){1-1} \cmidrule(r){2-6}  \cmidrule(r){7-11}  \cmidrule(r){8-16} 
    \textbf{HiT (-BA)} & 0.91  & 1.91  & 2.16  & 2.90  & 3.92  & 0.55  & 1.81  & 2.06  & 2.80  & 3.81  & 1.08  & 2.14  & 2.39  & 3.01  & 4.02  \\
    \textbf{HiT}   & 0.38  & 1.03  & 1.44  & 2.09  & 2.88  & 0.33  & 0.78  & 1.39  & 2.04  & 2.83  & 0.59  & 1.09  & 1.50  & 2.20  & 3.00  \\
    \bottomrule
    \end{tabular}%
    }
  \label{case_study}%
\end{table}%

\subsection{Case Studies on the Effectiveness of Behavior-aware Module}
To rigorously evaluate the impact of our proposed behavior-aware module, we conducted an in-depth analysis using the NGSIM dataset. This dataset was segmented into three distinct subsets—Low-level, Medium-level, and High-level aggressive driving behaviors—based on classifications generated by our fuzzy inference system within the behavior-aware module. These subsets account for 28.78\%, 56.37\%, and 14.85\% of the dataset, respectively, representing a wide spectrum of real-world behaviors.

We compared the performance of our HiT model against a version without the behavior-aware module (denoted as HiT (-BA)) and several leading baseline models. The results, presented in Table \ref{case_study}, demonstrate the substantial benefits of incorporating the behavior-aware module, particularly in the Low-level and High-level aggression subsets. For the Low-level subset, HiT outperforms other models by at least 7.7\% in short-term predictions and 10.3\% in long-term predictions. Similarly, for the High-level subset, HiT achieves a minimum improvement of 9.4\% over the baselines. In the more prevalent Medium-level subset, the performance gains are modest, suggesting that the module’s primary strength lies in handling extreme driving behaviors.

\begin{table}[t]
  \centering
  \caption{Characteristic analysis of driving behavior with NGSIM dataset. AVG. Speed: average speed; Avg. Acceleration: average acceleration;}
      \setlength{\tabcolsep}{2mm}
   \resizebox{0.6\linewidth}{!}{
    \begin{tabular}{c|ccc}
 \bottomrule 
    Type & AVG. Speed (km/h) & Avg. Acceleration (m/$s^2$) & Time Headway (s) \\
  \hline
    Low   & 32.5846 & 0.2321 & 3.451\\
    Medium & 43.3492 & 0.6387 & 2.624\\
   High  & 61.7271 & 2.41165 & 2.119\\
 \toprule
    \end{tabular}%
  \label{case_study_2}%
  }
\end{table}%

This differentiation in performance across aggression levels has significant practical implications. Early detection and accurate prediction of risky driving behaviors can enhance safety, and the behavior-aware module's ability to capture nuanced spatio-temporal interactions allows HiT to respond effectively to subtle environmental cues that might be overlooked by models lacking this capability.
To further analyze the impact of the behavior-aware module, we examined key driving metrics within each subset, focusing on vehicle speed, acceleration, and Time Headway. The mean and standard deviation values of these metrics, presented in Table \ref{case_study_2}, reveal that drivers exhibiting High-level aggressive behavior consistently engage in riskier practices, including higher speeds, more aggressive accelerations, and shorter Time Headway, compared to drivers in the Low-level subset. These patterns underscore the importance of the behavior-aware module in capturing and adapting to the real-time, non-linear dynamics of traffic interactions. Additionally, our analysis highlights the persistence of driving behaviors over time. By examining continuous driving data, HiT reliably classifies driver behaviors and predicts future actions, providing a solid foundation for proactive safety interventions. The contrast between HiT and HiT (-BA) further emphasizes the critical role of the behavior-aware module in enhancing predictive accuracy, particularly in high-risk scenarios involving aggressive driving or ensuring smooth navigation in less aggressive contexts. These case studies validate the effectiveness of the behavior-aware module and underscore its importance in improving the interpretability and reliability of behavior predictions.

\subsection{Ablation Studies}
To provide a detailed understanding of the contribution of each component in our model, We conducted comprehensive ablation studies for HiT. Detailed results and analyses of these ablation studies can be found in \textbf{Appendix \ref{Ablation Study}}.

\begin{figure}[t]
  \centering
\includegraphics[width=0.86\linewidth]{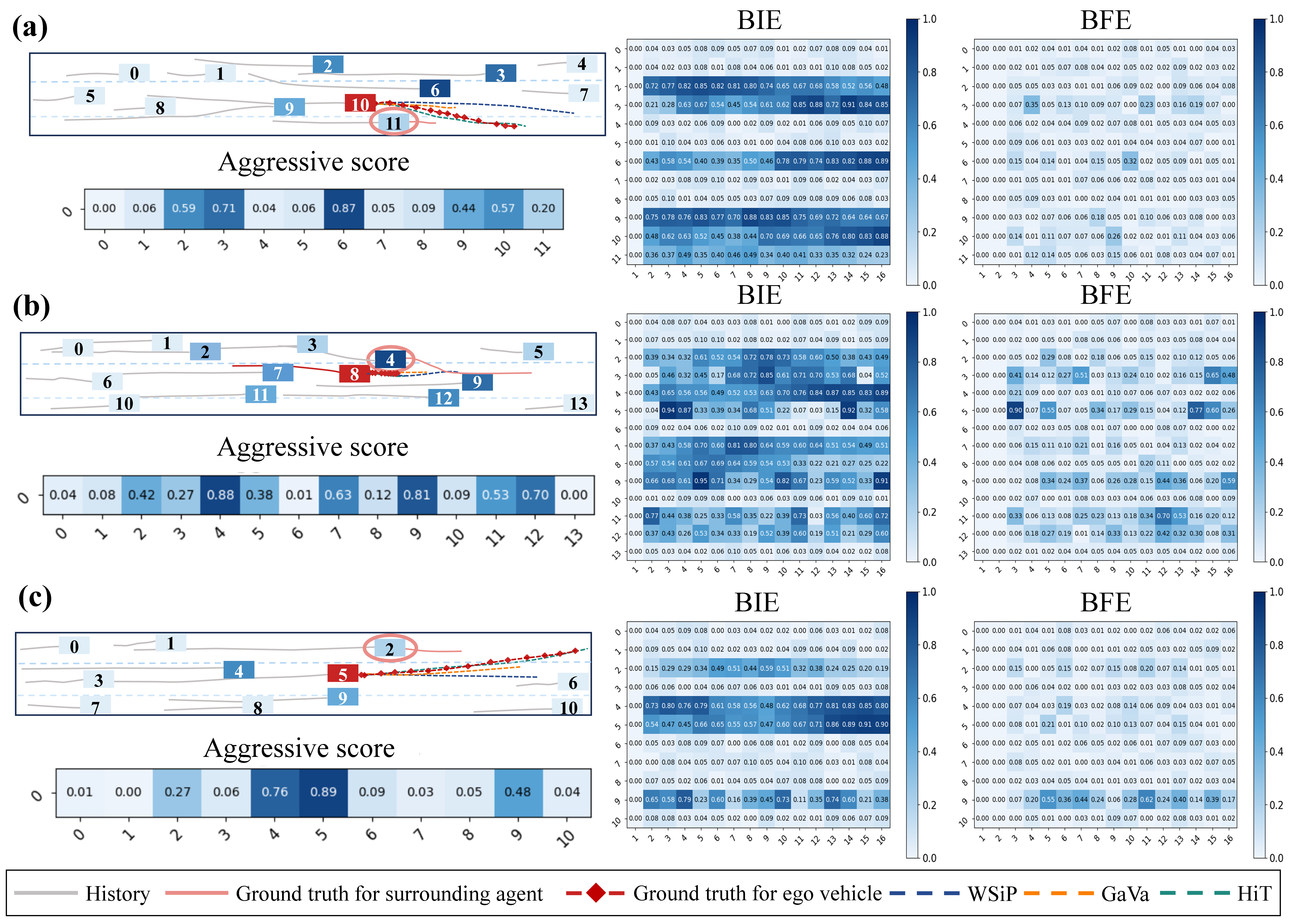} 
  \caption{Interpretability analysis of our proposed model HiT's trajectory predictions in complex traffic scenarios. The figure presents HiT’s predictive performance in three challenging driving scenarios: (a) a right-lane change maneuver with cooperative adjacent vehicles, (b) a right-lane change maneuver where an adjacent vehicle exhibits aggressive behavior, and (c) a lane-keeping scenario with varied interactions from surrounding vehicles. The ego vehicle is highlighted in red, whereas its surrounding agents are marked in blue. The heatmaps on the right illustrate the Behavior Intensity Estimate (BIE) and Behavior Fluctuation Estimate (BFE) for each scenario, showing how HiT evaluates the influence of surrounding vehicles based on their proximity and driving behavior. The results demonstrate HiT's ability to discern subtle social interactions and predict trajectories that closely align with the actual driving behavior, outperforming other models that do not incorporate behavior-awareness.}
  \label{fig_Behavior} 
\end{figure}

\subsection{Intuition and Interpretability Analysis}
\textbf{Intuition.} 
Our model is founded on the premise that driving behavior is inherently unique to each individual and remains relatively consistent throughout the driving process. This intrinsic behavior plays a pivotal role in shaping driving decisions and overall performance. Recognizing this, our approach considers the dynamic interplay within the driving environment, which includes both autonomous and human-driven vehicles. These vehicles' behaviors are influenced by a multitude of factors, including their relative position, speed, jerk profiles, behavioral tendencies, and historical driving patterns. To effectively capture and analyze these complexities, we have developed a comprehensive model that integrates behavioral, interactional, positional, and priority awareness through carefully crafted neural networks. This holistic design allows us to delve deeply into input trajectory data, enabling a thorough understanding of diverse driving scenarios and facilitating the accurate prediction of complex driving behaviors and trajectories. For instance, the trajectory of the ego vehicle is significantly influenced by the driving behaviors of surrounding vehicles. Successful lane merging, for example, is more likely when adjacent drivers exhibit conservative behaviors, such as slowing down to create space for the ego vehicle. However, failure to consider the potential actions of surrounding vehicles can introduce bias into the model’s predictions, leading to suboptimal decisions.

Notably, our proposed behavior-aware criteria provide a robust framework for enhancing the model’s environmental awareness. Specifically, we observe that aggressive drivers tend to exhibit sharp fluctuations in Behavior Intensity Estimate (BIE) and Behavior Fluctuation Estimate (BFE) over shorter time periods, characterized by rapid peaks and sudden increases in BIE. In contrast, conservative (Low-level aggressive) drivers show more gradual changes in these metrics, with a slower and steadier rise in BIE. The ability to identify and quantify these behaviors in real-time, made possible by our behavior-aware criteria, significantly enhances the model's predictive accuracy. By integrating these insights, our model can more effectively anticipate the movements of surrounding vehicles, leading to safer and more informed driving decisions. This approach ensures that the model not only predicts the most likely trajectory but also adapts to the behavior of other drivers, providing a more reliable and contextually aware driving experience.

\textbf{Interpretability Analysis.} 
The interpretability of predictive models in AD is crucial, not only for validating the model's effectiveness but also for ensuring trust in real-world applications. Fig. \ref{fig_Behavior} provides a comprehensive visual analysis of the HiT model's performance across different driving scenarios, showcasing its ability to anticipate driving behavior with remarkable accuracy.

In Fig. \ref{fig_Behavior} (a), the ego vehicle is engaged in a lane-changing maneuver. The behavior of the surrounding vehicles is critical in determining the success of this maneuver. Notably, a vehicle, highlighted with a red circle, demonstrates a cooperative driving style by adjusting its speed to allow the ego vehicle to merge smoothly. The corresponding heatmaps for Behavior Intensity Estimate (BIE) and Behavior Fluctuation Estimate (BFE) reflect this cooperation: the BIE heatmap shows a moderate but steady level of interaction, whereas the BFE remains low, indicating consistent and non-aggressive behavior. This scenario illustrates how HiT can recognize and leverage subtle cooperative behaviors in real-time, ensuring safe and efficient decision-making.

In contrast, Fig. \ref{fig_Behavior} (b) presents a more challenging situation where another vehicle, also marked with a red circle, exhibits aggressive behavior by accelerating to block the ego vehicle's lane change. This aggressive action is clearly captured in the BFE heatmap, which shows higher values indicating rapid changes in the surrounding vehicles' behavior. The BIE heatmap also reflects an increase in interaction intensity, highlighting the competitive nature of this driving scenario. HiT's ability to detect and predict such adversarial behavior is critical in urban environments, where unpredictable driver actions can lead to potential conflicts. The model's sensitivity to these nuances allows it to adjust the ego vehicle's trajectory proactively, avoiding unsafe situations.

Fig. \ref{fig_Behavior} (c) explores a scenario involving complex interactions among multiple vehicles in a congested traffic environment. Here, the ego vehicle must navigate around several other cars, each with varying degrees of influence on its trajectory. The BIE and BFE heatmaps reveal the intricate interplay between the vehicles. Some vehicles exert a strong influence despite being in non-adjacent lanes, a factor often overlooked by simpler models. The BIE values highlight the key interactions, whereas the BFE indicates fluctuations in behavior that might suggest changing traffic conditions or varying driver intentions. HiT's ability to account for these diverse influences, even from seemingly less relevant agents, is a testament to its advanced behavior-aware module.

What sets HiT apart is not just its predictive accuracy but its interpretive depth. By modeling driving behaviors in a way that mirrors human cognition, HiT provides a more holistic understanding of traffic dynamics. It doesn't merely predict where a vehicle will go; it predicts why it will go there, based on the observed behaviors of surrounding agents. This capability is crucial in scenarios where split-second decisions are necessary, and understanding the motivations behind other drivers' actions can mean the difference between a smooth lane change and a potential collision. Moreover, HiT's ability to capture the influence of non-adjacent vehicles—those in other lanes or further away—is a significant advancement over existing models like WSiP and GaVa, which may overlook these subtleties. HiT's nuanced approach allows it to predict not just the most likely trajectory, but the safest and most contextually appropriate one, considering the broader behavioral landscape. To sum up, HiT doesn't just predict; it observes, interprets, and decides like a human. By mirroring human decision-making, it offers a promising leap toward AD that's both accurate and reliable.

\section{Conclusion}\label{Conclusion} Predicting vehicle trajectories accurately is crucial for full AV development. We propose HiT, a human-like modular model with behavior-aware, interaction-aware, and multimodal decoding components. The behavior-aware module, based on dynamic geometric graph theory, offers flexibility by continuously representing driving behaviors, enabling real-time capture of vehicle dynamics. HiT outperforms state-of-the-art baselines on multiple datasets, demonstrating superior prediction accuracy and efficiency, even with limited training data. This model not only validates the effectiveness of behavior-aware methods but also shows its potential to reduce training data requirements, particularly in complex driving scenarios. HiT represents a significant advancement in AV trajectory prediction, offering a deeper understanding of driver behavior, vehicle interactions, and driving environment variability. Future work will focus on incorporating additional data sources, exploring new deep-learning techniques, and optimizing the model for real-world deployment. Techniques like model quantization and pruning will further improve scalability and computational efficiency, ensuring HiT’s practical application in fully autonomous driving systems.

\ACKNOWLEDGMENT{%
This research is supported by Science and Technology Development Fund of Macau SAR (File no. 0021/2022/ITP,  001/2024/SKL), Shenzhen-Hong Kong-Macau Science and Technology Program Category C (SGDX20230821095159012), State Key Lab of Intelligent Transportation System (2024-B001), Jiangsu Provincial Science and Technology Program (BZ2024055), and University of Macau (SRG2023-00037-IOTSC).
Please ask Dr. Zhenning Li (zhenningli@um.edu.mo) for correspondence.
}

\bibliographystyle{informs2014trsc}
\bibliography{cas-refs}

\newpage
% \bibliographystyle{cas-refs}

% Appendix here
% Options are (1) APPENDIX (with or without general title) or 
%             (2) APPENDICES (if it has more than one unrelated sections)
% Outcomment the appropriate case if necessary
%
% \begin{APPENDIX}{<Title of the Appendix>}
% \end{APPENDIX}
%
%   or 
%
\begin{APPENDIX}{Towards Human-Like Trajectory Prediction for Autonomous Driving: A Behavior-Centric Approach}

\section{Multimodal Probabilistic Maneuver Prediction}\label{Multimodal Probabilistic}
In our proposed model, future trajectories are predicted hierarchically within a Bayesian framework, operating at two levels. First, at each time step, the probability of different maneuvers $\bm{M}$ for the ego vehicle is calculated. Next, based on each maneuver, detailed trajectories are generated within a predefined distribution. As shown in Fig. \ref{manuver}, to account for the driver's actions during driving, possible maneuvers are decomposed into two subcategories: positioning-wise sub-maneuvers ${M}_{p}$ and velocity-wise sub-maneuvers ${M}_{v}$. Positioning-wise maneuvers include left lane change, right lane change, and lane keeping, whereas velocity-wise maneuvers include accelerating, braking, and speed maintenance.

\begin{figure}[htbp]
  \centering
\includegraphics[width=0.8\linewidth]{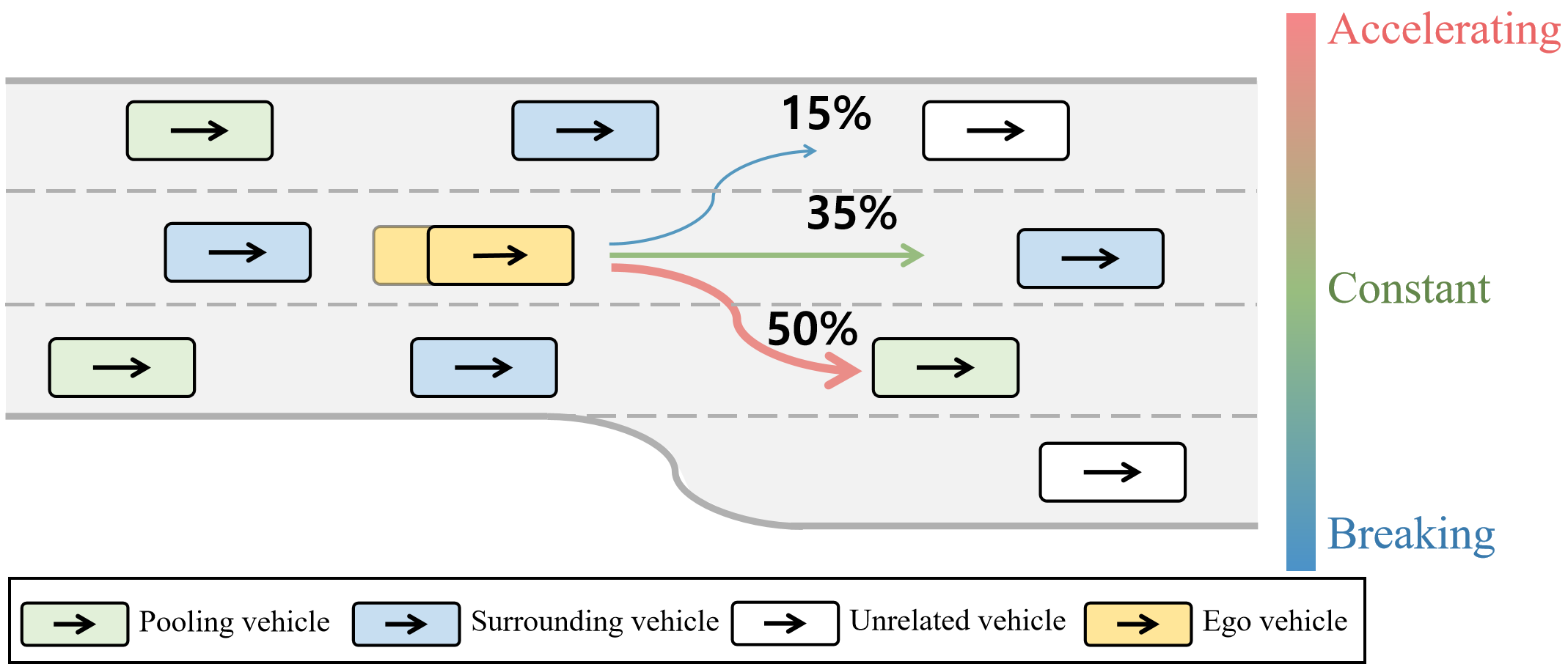} 
  \caption{Multimodal maneuver prediction framework with corresponding probability outputs. Regarding maneuver uncertainty, we categorize the possible maneuvers based on the characteristics of the driver's actions and enable the model to predict based on these potential maneuvers. These maneuvers include speed-related maneuvers: accelerating, braking, and maintaining speed, and position-related maneuvers: left lane change, right lane change, and lane keeping.}
  \label{manuver} 
\end{figure}

The framework further refines the prediction by generating detailed trajectories conditioned on each maneuver within a specific distributional form. At each time step \(t\), the task extends to estimate the distribution of plausible future trajectories:
\begin{equation}
    \bm{P}\left(\bm{Y}_{0}^{t:t+t_{f}} \mid \bm{M}, \bm{X}_{0:n}^{t-t_{h}:t}\right)
\end{equation}
Given the estimated maneuvers \(\bm{M}\), the probability distribution of multimodal predictions \(\bm{Y}_{0}^{t:t+t_{f}}\) is parameterized as a bivariate Gaussian distribution with estimable parameters \(\bm{\Omega}\):
\begin{equation}
\bm{P}\left(\bm{Y}_{0}^{t:t+t_{f}} \mid \bm{M}, \bm{X}_{0:n}^{t-t_{h}:t}\right) = \bm{P}_{\bm{\Omega}} (\bm{Y}_{0}^{t:t+t_{f}} \mid \bm{M}, \bm{X}_{0:n}^{t-t_{h}:t}) = \bm{N}(\bm{Y}_{0}^{t:t+t_{f}}|\bar{\mu}(\bm{M}, \bm{X}_{0:n}^{t-t_{h}:t}),\Sigma(\bm{M}, \bm{X}_{0:n}^{t-t_{h}:t}))
\end{equation}
Here, \(\bm{\Omega}=\left[\Omega^{t+1}, \ldots,\Omega^{t+t_{f}}\right]\), and \(\Omega^{t}=[\bar{\mu}_{t, \rho}, \bar{\mu}_{t, \theta}, \sigma_{t, \rho}, \sigma_{t, \theta}, \eta_{t}]\) represent the mean, variance, and correlation coefficient of the vehicle's distance and orientation, reflecting the uncertainty of the prediction at time \(t\). 

The predicted means $\bar{\mu}(\bm{M}, \bm{X}_{0:n}^{t-t_{h}:t})$ and covariance matrix $\Sigma(\bm{M}, \bm{X}_{0:n}^{t-t_{h}:t})$ are expressed as follows: 
\begin{equation}
\bar{\mu}(\bm{M}, \bm{X}_{0:n}^{t-t_{h}:t}) = [\bar{\mu}_{t+1, \rho}, \bar{\mu}_{t+1, \theta}, \ldots, \bar{\mu}_{t+t_{f}, \rho}, \bar{\mu}_{t+t_{f}, \theta}],
\end{equation}
\begin{equation}
\begin{aligned}
\Sigma(\bm{M}, \bm{X}_{0:n}^{t-t_{h}:t}) 
 = \begin{pmatrix}
\sigma_{t+1, \rho}^2 & 0 & 0 & \ldots & 0 & 0 & 0 \\
0 & \sigma_{t+1, \theta}^2 & 0 & \ldots & 0 & 0 & 0 \\
0 & 0 & \eta_{t+1} & \ldots & 0 & 0 & 0 \\
\vdots & \vdots & \vdots & \ddots & \vdots & \vdots & \vdots \\
0 & 0 & 0 & \ldots & \sigma_{t+t_{f}, \rho}^2 & 0 & 0 \\
0 & 0 & 0 & \ldots & 0 & \sigma_{t+t_{f}, \theta}^2 & 0 \\
0 & 0 & 0 & \ldots & 0 & 0 & \eta_{t+t_{f}}
\end{pmatrix}
\end{aligned}
\end{equation}
The multimodal predictions are then formulated as a Gaussian Mixture Model (GMM), where the overall probability distribution of the future trajectory $\bm{Y}_{0}^{t:t+t_{f}}$ is computed as:
\begin{equation}
\begin{aligned}
&\bm{P}\left(\bm{Y}_{0}^{t:t+t_{f}} \mid \bm{M}, \bm{X}_{0:n}^{t-t_{h}:t}\right)\\ &= \sum_{\forall i} \bm{P}\left({M}_{i} \mid \bm{X}_{0:n}^{t-t_{h}:t}\right) \bm{P}_{\bm{\Omega}}\left(\bm{Y}_{0}^{t+1:t+t_{f}} \mid {M}_{i},\bm{X}_{0:n}^{t-t_{h}:t}\right) \\
 &= \sum_{\forall i} \bm{P}\left({M}_{i} \mid \bm{X}_{0:n}^{t-t_{h}:t}\right) \frac{1}{\sqrt{(2\pi)^{d}|\Sigma_{i}(\bm{M}, \bm{X}_{0:n}^{t-t_{h}:t})|}} \exp \{\\& -\frac{1}{2} (\bm{Y}_{0}^{t:t+t_{f}} - \mu_{i}(\bm{M}, \bm{X}_{0:n}^{t-t_{h}:t}))^T \Sigma_{i}(\bm{M}, \bm{X}_{0:n}^{t-t_{h}:t})^{-1} (\bm{Y}_{0}^{t:t+t_{f}} - \mu_{i}(\bm{M}, \bm{X}_{0:n}^{t-t_{h}:t})) \}
\end{aligned}
\end{equation}
where \(M_{i}\) is the \(i\)th element in \(\bm{M}\). This multimodal structure provides not only multiple trajectory predictions but also quantifies the confidence associated with each, supporting more robust decision-making in the face of prediction uncertainty.

\section{Proof for Theorem 1}\label{Proof_ICC_1}
\begin{myproof}
Let \( G^t = (V^t, E^t) \) and \( G'^t = (V^t, E'^t) \) be two DGGs at time \( t \), where the node sets are identical, but the edge sets differ. The respective adjacency matrices for these graphs are denoted by \( A^t \) and \( A'^t \). The only difference between \( A^t \) and \( A'^t \) is the connection status between a pair of nodes \( v_k^t \) and \( v_\ell^t \), such that \( A'^t(k, \ell) = 1 - A^t(k, \ell) \) and \( A'^t(\ell, k) = 1 - A^t(\ell, k) \).
The degree centrality \( d_i(G^t) \) of node \( v_i^t \) is defined as the sum of the connections between \( v_i^t \) and its adjacent nodes, given by\(
d_i(G^t) = \sum_{v_j^t \in N_i^t} A^t(i, j),
\), where \( N_i^t \) denotes the set of nodes adjacent to \( v_i^t \) in \( G^t \). 
Consider two arbitrary nodes \( v_i^t \) and \( v_j^t \) in the set \( V^t \setminus \{v_k^t, v_\ell^t\} \).
Since the only difference between \( G^t \) and \( G'^t \) is the presence or absence of the edge \( (v_k^t, v_\ell^t) \), and \( v_i^t \) and \( v_j^t \) are not involved in this edge, their respective adjacency sets \( N_i^t \) and \( N_j^t \) remain identical in both graphs. Consequently, for any \( v_m^t \in N_i^t \) and \( v_n^t \in N_j^t \), we have \( A^t(i, m) = A'^t(i, m) \) and \( A^t(j, n) = A'^t(j, n) \), leading directly to \(
d_i(G^t) = d_i(G'^t)\) and \(d_j(G^t) = d_j(G'^t)\).
Given that \( d_i(G^t) \geq d_j(G^t) \) implies \( d_i(G'^t) \geq d_j(G'^t) \), we can conclude that degree centrality satisfies the IIC property.
 This implies that the ranking of degree centrality between nodes not directly involved in the edge modification remains unaffected by the addition or removal of that specific edge.
\end{myproof}

\section{Proof for Theorem 2}  \label{Proof_ICC_2}
\begin{myproof}
Consider two dynamic geometric graphs at time \( t \), denoted by \( G^t = (V^t, E^t) \) and \( G'^t = (V^t, E'^t) \), where the node sets are identical, but the edge sets differ. Assume that there exists an edge \( (v_k^t, v_\ell^t) \) in \( G^t \), which is removed in \( G^{'t} \). The corresponding kernel-weighted multi-feature adjacency matrices, \( \bar{A}^t \) and \( \bar{A}^{'t} \), thus satisfy:
$\bar{a}_{k\ell}^t \neq \bar{a}_{k\ell}^{'t}$ and $ \bar{a}_{\ell k}^t \neq \bar{a}_{\ell k}^{'t}$, where 
resulting in a difference between the kernel-weighted multi-feature Laplacian matrices \( \bar{L}^t \) and \( \bar{L}^{'t}\).

According to the definition of the dynamic degree centrality,  the dynamic degree centrality \( \mathcal{J}_{i}^{t}(D) \) of the $i$-th vehicle \( v_i^t \) at time $t$ can be given as follows:
\begin{equation}
\mathcal{J}_{i}^{t}(D) = 
x_i^t(\varepsilon) + \varepsilon \sum_{v_j^t \in N_i^t} \bar{a}_{ij}^t \left[x_i^t(\varepsilon) - x_j^t(\varepsilon)\right]
\end{equation}
where \( \mathbf{x}^t(\varepsilon) \) satisfies the equation:
\begin{equation}
(I + \varepsilon \bar{L}^t) \mathbf{x}^t(\varepsilon) = \mathbf{d}^t,
\end{equation}
For the DGGs \( G^t \) and \( G'^t \), the corresponding centralities are denoted by \( \mathcal{J}_{i}^{t}(D) \) and \( \mathcal{J}_{i}^{t}(D') \), respectively, with their difference arising from the variations between \( \bar{L}^t \) and \( \bar{L}'^t \). Given that \( \bar{L}^t \neq \bar{L}'^t \), it follows the following equation:
\begin{equation}
\mathbf{x}^t(\varepsilon) \neq \mathbf{x}'^t(\varepsilon)
\end{equation}
which subsequently implies:
\begin{equation}
\mathcal{J}_{i}^{t}(D) \neq \mathcal{J}_{i}^{t}(D')
\end{equation}
Obviously, if \( \mathcal{J}_{i}^{t}(D) \geq \mathcal{J}_{j}^{t}(D) \), this relation may not hold in \( G^{'t} \), i.e., \( \mathcal{J}_{i}^{t}(D') \geq \mathcal{J}_{j}^{t}(D') \) may not be satisfied. Even though \( v_i^t \) and \( v_j^t \) are not directly involved with the edge \( (v_k^t, v_\ell^t) \), the indirect connections influenced by the change in \( \bar{L}^t \) and \( \bar{L}^{'t} \) could alter the relative centralities \( \mathcal{J}_{i}^{t}(D) \) and \( \mathcal{J}_{j}^{t}(D) \). Formally,
\begin{equation}
    \mathcal{J}_{i}^{t}(D) \geq \mathcal{J}_{j}^{t}(D) \quad \nRightarrow \quad \mathcal{J}_{i}^{t}(D') \geq \mathcal{J}_{j}^{t}(D')
\end{equation}
which proves that the dynamic degree centrality does not satisfy the IIC criterion for any \( \varepsilon > 0 \).
\end{myproof}

\section{Member Functions}  \label{FIS_1}

The Low (L)/Medium (M)/High (H) membership functions for BIE are defined as follows:
% Low (L):
\begin{small}
   \begin{equation}
  \mu_{\text{L}}^{\text{BIE}}(x) =
  \begin{cases} 
  1 - \frac{x}{0.3}, & 0 \leq x < 0.3 \\
  0, & x \geq 0.3
  \end{cases}
  ,\;
  \mu_{\text{M}}^{\text{BIE}}(x) =
  \begin{cases} 
  \frac{x-0.2}{0.5-0.2}, & 0.2 < x \leq 0.5 \\
  \frac{0.8-x}{0.8-0.5}, & 0.5 < x \leq 0.8 \\
  0, & \text{otherwise}
  \end{cases}
 ,\;
    \mu_{\text{H}}^{\text{BIE}}(x) =
  \begin{cases} 
  0, & x < 0.7 \\
  \frac{x-0.7}{1-0.7}, & 0.7 \leq x < 1 \\
  1, & x = 1
  \end{cases}
  \end{equation}
  \end{small}

Correspondingly, the following are the   Low (L)/Medium (M)/High (H) member functions for BFE:
% Low (L):
\begin{small}
\begin{equation}
  \mu_{\text{L}}^{\text{BFE}}(x) =
  \begin{cases} 
  1 - \frac{x}{0.35}, & 0 \leq x < 0.35 \\
  0, & x \geq 0.35
  \end{cases},\;
    \mu_{\text{M}}^{\text{BFE}}(x) =
  \begin{cases} 
  \frac{x-0.3}{0.5-0.3}, & 0.3 < x \leq 0.5 \\
  \frac{0.75-x}{0.75-0.5}, & 0.5 < x \leq 0.75 \\
  0, & \text{otherwise}
  \end{cases},\;
    \mu_{\text{H}}^{\text{BFE}}(x) =
  \begin{cases} 
  0, & x < 0.65 \\
  \frac{x-0.65}{1-0.65}, & 0.65 \leq x < 1 \\
  1, & x = 1
  \end{cases}
 \end{equation}
\end{small}

Additionally, the  Low (L)/Medium (M)/High (H) member functions of output aggressive scores can be defined as follows:
% Low (L):
\begin{small}
 \begin{equation}
\mu_{\text{L}}^{\text{Score}}(s) = 
\begin{cases} 
1, & 0 \leq s < 0.2 \\
\frac{0.4 - s}{0.4 - 0.2}, & 0.2 \leq s \leq 0.4 \\
0, & s > 0.4 
\end{cases}, \;
\mu_{\text{M}}^{\text{Score}}(s) = 
\begin{cases} 
0, & s < 0.3 \\
\frac{s - 0.3}{0.5 - 0.3}, & 0.3 \leq s < 0.5 \\
1, & s = 0.5 \\
\frac{0.7 - s}{0.7 - 0.5}, & 0.5 < s \leq 0.7 \\
0, & s > 0.7 
\end{cases},\;
\mu_{\text{H}}^{\text{Score}}(s) = 
\begin{cases} 
0, & s < 0.6 \\
\frac{s - 0.6}{0.8 - 0.6}, & 0.6 \leq s < 0.8 \\
1, & s = 0.8 \\
\frac{1.0 - s}{1.0 - 0.8}, & 0.8 < s \leq 1.0 \\
0, & s > 1.0 
\end{cases}
 \end{equation}
 \end{small}

% \section{Proof for Theorem 3}  \label{FIS_2}

\section{Interaction-aware Module}\label{Interaction-aware Module}

\textbf{GRU Encoder.} A stack of GRU layers is used in this encoder to capture the temporal relationships within the input observations \( \bm{X} \), transforming them into a sequence of tokens $\mathbf{O}_{\text{tok}}$. 

\textbf{Positional Encoding.} Following this, we employ positional encoding to incorporate the spatial insights provided by the polar coordinate system. Our implementation of positional encoding is inspired by the approach of Vaswani et al. \citep{vaswani2017attention}, using sine and cosine functions of varying frequencies. This method enables the model to encode positional information in a continuous and differentiable manner, thereby enhancing the efficiency and effectiveness of the training process. Specifically, each positional encoding dimension corresponds to a sinusoid from $2 \pi$ to $10000 \cdot 2 \pi$ with wavelengths that form a geometric progression. This allows the model to learn to pay more attention to positions and use this information to better determine the importance of surrounding vehicles to the ego vehicle. Formally, the positional encoding ${\textit{PE}}$ can be  defined as follows:
\begin{equation}\label{eq.14-1}
    \begin{aligned}
    \textit{PE}_{(\mathbf{O}_{\text{tok}}, 2\kappa)} &=\sin \left(\mathbf{O}_{\text{tok}}/10000^{2\kappa/d_{\textit{model}}}\right) \\
    \textit{PE}_{(\mathbf{O}_{\text{tok}}, 2\kappa+1)} &=\cos \left(\mathbf{O}_{\text{tok}} /10000^{2\kappa /d_{\textit{model}}}\right)
    \end{aligned}
\end{equation}
where $d_{\textit{model}}$ is the dimension, $\mathbf{O}_{\text{tok}}$ is each position information embedded by the GRU encoder with $\kappa$ denotes the dimension of $\mathbf{O}_{\text{tok}}$. The output of positional encoding is position vectors \( \mathbf{O}_{\text{pos}} \).

\textbf{Interaction Encoder.}
This encoder is crafted to further capture the spatio-temporal interactions for the position vectors \( \mathbf{O}_{\text{pos}} \). We introduce a lightweight transformer-based framework to extract interaction features \( \mathbf{\bar{F}}_i \). Specifically, we utilize a multi-head attention mechanism where the features of the ego vehicle serve as the query \( \mathbf{\bar{Q}} \in \mathbb{R}^{1 \times d_{\textit{model}}} \), and the features of all other vehicles within the pooling range are used as the keys \( \mathbf{\bar{K}} \in \mathbb{R}^{n \times d_{\textit{model}}} \) and values \( \mathbf{\bar{V}} \in \mathbb{R}^{n \times d_{\textit{model}}} \). This design enables the model to effectively weigh the influence of surrounding vehicles, capturing the intricate interactions within the traffic environment. Formally, 
\begin{equation}\label{}
    \begin{aligned}
    \mathbf{\bar{F}^\prime}_i &=\phi_{\text{Concat}}(\text{head}_1, \text{head}_2, \cdots, \text{head}_{\bar{h}})\mathbf{\bar{W}}^O \\
     \textit{where} \quad \text{head}_i&= \phi_{\text{softmax}}\left(\frac{\mathbf{(\bar{Q}}_i +\mathbf{\bar{W}}_i^Q \mathbf{O}_{(\text{tok}, i)}^{Q_i}) ({\mathbf{\bar{K}}_i +\mathbf{\bar{W}}_i^K \mathbf{O}_{(\text{tok}, i)}^{K_i}})^\top}{\sqrt{d_k}}\right)\mathbf{\bar{V}}_i
    \end{aligned}
\end{equation}
where $\phi_{\text{softmax}}$ denotes the softmax activaiton function and  $d_k = d_{\textit{model}} / \bar{h}$, $\bar{h}$ is the number of heads. $ \mathbf{\bar{Q}}_i \in \mathbb{R}^{1 \times d_{k}}$, $ \mathbf{\bar{K}}_i \in \mathbb{R}^{n \times d_{k}}$, and $ \mathbf{\bar{V}}_i \in \mathbb{R}^{n \times d_{k}}$ represent the features of head $i$. $\mathbf{\bar{W}}_i^K \in \mathbb{R}^{n \times n} $, $\mathbf{\bar{W}}_i^Q \in \mathbb{R}^{1 \times 1}$, and $\mathbf{\bar{W}}^O \in \mathbb{R}^{d_{k} \times d_{\text{model}}}$ are the learnable weight matrices. 
$\mathbf{O}_{(\text{tok}, i)}^{Q_i} \in \mathbb{R}^{n \times d_{k}}$ and 
$\mathbf{O}_{(\text{tok}, i)}^{K_i} \in \mathbb{R}^{1 \times d_{k}}$ are the output of positional encoding  for head $i$. $\mathbf{\bar{F}^\prime}_i \in \mathbb{R}^{\bar{h} \times d_{\textit{model}}}$ is the output of attention mechanism.

In addition, we introduce a dynamically gated Feed-Forward Network (FFN) to enhance the model's spatial awareness. The key advantage of the gating mechanism is that it generates a weighting factor, which dynamically adjusts the degree of fusion between input features and spatial information. This dynamically spatial-enhanced FFN design enables the model to more flexibly integrate these inputs, adapting the weights of both based on the context. This capability is particularly beneficial in capturing spatial-dependent relationships during complex interactions between the target vehicle and surrounding vehicles, especially in dynamic traffic scenarios. We formulate this process as follows:
\begin{equation}
    \begin{aligned}
\mathbf{\bar{F}}_i =\max \left(0, g(\mathbf{\bar{F}^\prime}_i, \mathbf{O}_{\text{pos}}) \cdot (\mathbf{\bar{F}^\prime}_i \mathbf{\bar{W}}_1+\bar{b}_1)^\top\right) \mathbf{\bar{W}}_2+\bar{b}_2 \\
     \textit{where} \quad g(\mathbf{\bar{F}^\prime}_i, \mathbf{O}_{\text{pos}}) =\phi_{sig}\left(\mathbf{\bar{W}}_g[\mathbf{\bar{F}^\prime}_i ; \mathbf{O}_{\text{pos}}]+\bar{b}_g\right)
    \end{aligned} 
\end{equation}
where $\phi_{sig}$ is the sigmoid activation function. $[\cdot]$ denotes the concat operations. $ \mathbf{O}_{\text{pos}} \in \mathbb{R}^{(n+1) \times d_{\textit{model}}}$ , $\mathbf{\bar{W}}_g \in \mathbb{R}^{d_{\textit{model}} \times (\bar{h}+n+1)}$, $\mathbf{\bar{W}}_1 \in \mathbb{R}^{d_{\textit{model}} \times  d_{\textit{model}}  }$, and $\mathbf{\bar{W}}_2 \in \mathbb{R}^{\bar{h} \times  (n+1)  }$ denote the learnable weights. $\bar{b}_1$, $\bar{b}_2$, and $\bar{b}_g$ are the bias terms. $\mathbf{\bar{F}}_i \in \mathbb{R}^{d_{\textit{model}} \times  (n+1)}$.
Given the interaction feature $\mathbf{\bar{F}}_i$, we apply a layer of MLP to obtain the final output interaction vector $\mathbf{O}_{int} = \phi_{\textit{MLP}}(\mathbf{\bar{F}}_i^\top) \in \mathbb{R}^{(n+1) \times d_\textit{model}}$.

Overall, the interaction vector $\mathbf{O}_{\text{int}}$ is generated by the interaction-aware module to model the interactions between the ego vehicle and its surrounding agents. The module begins with a GRU encoder, which captures the spatial dynamics of each vehicle from the historical observations $\bm{X}$. The GRU output tokens, $\mathbf{O}_{\text{tok}}$, are then enhanced with positional encoding to incorporate temporal dependencies, resulting in position vectors $\mathbf{O}_{\text{pos}}$. The interaction encoder then progressively captures the spatio-temporal interactions, dynamically weighting the hidden states of the position vectors to produce the high-level interaction vector $\mathbf{O}_{\text{int}}$.

\section{Loss Functions}\label{Loss}
In alignment with the multi-task learning framework proposed by Kendall et al. \citep{kendall2018multi}, the overall loss function for other datasets, denoted as \(\mathcal{L}_{\textit{overall}}\), integrates both the maneuver loss \(\mathcal{L}_\textit{man}\) and the trajectory loss \(\mathcal{L}_\textit{tra}\). To appropriately balance these loss functions, we apply a Gaussian probability estimate incorporating homoscedastic uncertainty. The multitask probability over the outputs is formulated:
\begin{equation}
\begin{aligned}
p\left(\mathcal{L}_\textit{tra}, \mathcal{L}_\textit{man} \mid {f}^{\mathbf{W}^{\circ}}(\bm{X})\right)=p\left(\mathcal{L}_\textit{tra} \mid {f}^{\mathbf{W}^{\circ}}(\bm{X})\right) p\left(\mathcal{L}_\textit{man} \mid {f}^{\mathbf{W}^{\circ}}(\bm{X})\right)
\end{aligned}
\end{equation}

Here, ${f}^{\mathbf{W}}(\bm{X})= \bm{Y}_{0}^{t:t+t_{f}}$ denotes the output of our model with weights $\mathbf{W}^{\circ}$ for input $\bm{X} = \bm{X}_{0:n}^{t-t_h:t} $. 
We aim to maximize the Log-Likelihood and its parameters, including the weight $\mathbf{W}^{\circ}$ and the noise parameters $\sigma_{\textit{tra}}$ and $\sigma_{\textit{man}}$, thereby minimizing loss $\mathcal{L}_{\textit{overall}}$ for the output. Formally,
\begin{equation}
\begin{aligned}
\mathcal{L}_{\textit{overall}}&=-\log p\left(\mathcal{L}_\textit{tra}, \mathcal{L}_\textit{man} \mid {f}^{\mathbf{W}^{\circ}}(\mathbf{X})\right) \\
& \propto \frac{1}{2 \sigma_{\textit{tra}}^2}\left\|\mathcal{L}_\textit{tra}-{f}^{\mathbf{W}^{\circ}}(\mathbf{X})\right\|^2+\frac{1}{2 \sigma_{\textit{man}}^2}\left\|\mathcal{L}_\textit{man}-{f}^{\mathbf{W}^{\circ}}(\mathbf{X})\right\|^2+\log (\sigma_{\textit{tra}} \sigma_{\textit{man}}) \\
& =\frac{1}{2 \sigma_{\textit{tra}}^2} \mathcal{L}_\textit{tra}(\mathbf{W}^{\circ})+\frac{1}{2 \sigma_{\textit{man}}^2} \mathcal{L}_\textit{man}(\mathbf{W}^{\circ})+\log (\sigma_{\textit{tra}} \sigma_{\textit{man}})
\end{aligned}
\end{equation}

\section{Experimental Results}
\subsection{Performance Evaluation on the \textit{Overall} Test Set.} \label{Overall Experimental Results}
\textit{NGSIM:} We utilized the NGSIM dataset to evaluate the HiT model's prediction capabilities in American highway scenarios. As shown in Table \ref{tab:addlabel}, HiT achieved the best performance across all prediction horizons. Notably, the advantage of our model became increasingly pronounced as the prediction horizon extended. At the 5-second prediction horizon, HiT surpassed the leading model, GaVa \citep{liao2024human}, by 0.26 meters. Additionally, in the 4-second prediction horizon, we observed a 7.1\% improvement. These results underscore the HiT's exceptional ability to comprehend and predict behavior in highway scenarios.

\textit{HighD:} We conducted comparative experiments using the HighD dataset to validate the superiority of our model in German highway scenarios. As shown in Table \ref{tab:addlabel}, our HiT model consistently maintained top performance in long-term prediction horizons, achieving improvements of 22\% and 15.3\% at the 4-second and 5-second prediction horizons, respectively. Although our model did not achieve the best performance in the short-term prediction horizons (1 to 3 seconds), the differences were minimal. For instance, within the 3-second prediction horizon, our model trails iNATran by only 0.01 meters. However, in the more challenging long-term predictions, HiT demonstrates a significant advantage. Moreover, when considering the overall performance across the entire prediction period, HiT achieves the lowest average RMSE, outperforming the second-best baseline, Wave, by 28.2\%. These demonstrate HiT is well-adapted to the highway scenarios.

\textit{RounD:} We evaluated the performance of our model and various baseline models using the RounD dataset, with the comparative results presented in Table \ref{tab:addlabel}. It is evident that our model outperformed all baselines across every prediction horizon. Notably, at the 1-second prediction horizon, our model achieved a remarkable 40\% improvement. Additionally, in terms of the overall Average RMSE metric, HiT demonstrated a 22.6\% enhancement. These results indicate the superior performance of HiT in roundabout scenes.

\textit{MoCAD++:} To assess the predictive capabilities of our model in urban and campus scenarios, we tested various SOTA models on the MoCAD++ dataset, with the comparative results summarized in Table \ref{tab:addlabel}. Compared to the previous SOTA model, GaVa \citep{liao2024human}, our proposed model demonstrated improvements across all prediction horizons. Notably, in the 1-second short-term prediction horizon, the increase reached 45.8\%. Additionally, our model achieved a 9.2\% improvement in the overall metric of Average RMSE. Interestingly, in the right-hand-drive dynamic of the MoCAD++ dataset, the performance of these SOTA baselines did not match their results on other datasets, likely because of the unique traffic regulations and complex driving environments associated with this dataset. However, HiT's advantage in this challenging scenario became even more pronounced, further underscoring the underscore the adaptability and robustness of our proposed HiT model in urban and campus traffic environments.

\textit{ApolloScape:} As shown in Table \ref{tab:apollo}, the ApolloScape dataset emphasizes the overall performance of models in heterogeneous traffic flows, making WSFDE and WSADE the primary metrics of interest. In terms of WSFDE, which is a comprehensive measure of the final destination accuracy across vehicles, pedestrians, and bicycles, our model achieves a 10.1\% improvement over the previous best model. For WSADE, our model also demonstrates a 2.6\% improvement. We then proceed to discuss the superiority of our model across different categories. For vehicles, our model surpasses the existing baselines by 9.2\% and 9.6\% in ADE and FDE metrics, respectively. Regarding pedestrians, our model exceeds the baselines by 3.6\% and 4.7\% in ADE and FDE metrics, respectively. For bicycles, our model outperforms the baselines by 9.2\% in ADE and 8.7\% in FDE. These results collectively demonstrate the robustness and predictive capability of our proposed model.

To evaluate the scalability and efficiency of HiT, we trained the model on a reduced 25\% subset of the training data and assessed its performance on both the overall test set and the maneuver-based test set. As presented in Table \ref{Table1} and Table \ref{Table_manuver}, HiT achieves significantly lower RMSE values than most baseline models, even with limited training data. This demonstrates HiT's ability to reduce data requirements for training autonomous vehicles, especially in challenging scenarios.

\subsection{Performance Evaluation on Maneuver-based Test Set} \label{Maneuver}  
Table \ref{Table_manuver} presents the test performance of HiT and other baselines on the Maneuver-based Test Set. The results indicate that the HiT model outperformed all baselines in terms of RMSE, achieving improvements of 16.9\%, 14.5\%, 13.2\%, and 21.9\% across the \textit{keep}, \textit{merge}, \textit{left}, and \textit{right} subsets, respectively. When evaluating the 5-second prediction horizon, our model demonstrated a 14.1\% improvement over the previously best-performing GaVa model \citep{liao2024human} in the \textit{right} subset. Additionally, our model exhibited significant gains in the \textit{right} test subsets, further highlighting its adaptability across diverse driving scenarios.

\begin{table}[htbp]
  \centering
  \caption{Evaluation results for the proposed model and the baselines in the maneuver-base test set for the NGSIM dataset. \textbf{Bold} and \underline{underlined} values represent the best and second-best performance in each category.}%dataset over a 5-second forecast horizon
  \setlength{\tabcolsep}{2mm}
   \resizebox{0.8\linewidth}{!}{
    \begin{tabular}{ccccccccccc}
    \toprule
    Dataset & \multicolumn{5}{c}{\textit{keep}}     & \multicolumn{5}{c}{\textit{merge}} \\
    \cmidrule(r){1-6} \cmidrule(r){7-11} 
    \multirow{2}[1]{*}{Model} & \multicolumn{5}{c}{Horizon (s)}       & \multicolumn{5}{c}{Horizon (s)} \\
\cmidrule(r){2-6} \cmidrule(r){7-11}          & 1     & 2     & 3     & 4     & 5     & 1     & 2     & 3     & 4     & 5 \\
    S-LSTM \citep{alahi2016social} & 0.35  & 1.01  & 1.81  & 2.82  & 4.15  & 0.81  & 1.31  & 2.51  & 4.01  & 5.78 \\
    S-GAN \citep{gupta2018social} & 0.36  & 1.01  & 1.81  & 2.83  & 4.15  & 0.71  & 1.32  & 2.53  & 4.11  & 5.97 \\
    CS-LSTM \citep{deo2018convolutional} & 0.34  & 0.98  & 1.75  & 2.77  & 4.06  & 0.61  & 1.34  & 2.58  & 4.12  & 5.94 \\
    STDAN \citep{chen2022intention}& \underline{0.28}  & 0.85  & {1.52}  & 2.53  & {3.49}  & \underline{0.28}  & 1.19  & {2.21}  & 3.67  & {4.95} \\
     WSiP \citep{wang2023wsip} & 0.32  & 0.89  & 1.58  & {2.51}  & 3.59  & 0.40  & {1.18}  & 2.41  & 3.72  & 5.16 \\

    GaVa \citep{liao2024human} &  \underline{0.28}   & \underline{0.77}    & 1.49     & \underline{2.01}     & 2.86     & 0.33     & \underline{0.84}     & \underline{1.63}     & 2.45     & \underline{3.34} \\ 
    \cmidrule(r){1-1} \cmidrule(r){2-6}  \cmidrule(r){7-11}
    \textbf{HiT (25\%)} &  0.32  & 0.82    & \underline{1.32}     &   2.07   & \underline{2.83}     & 0.41     & 0.95     & 1.68     & \underline{2.28}    & 3.41 \\ 
    
    \textbf{HiT} & \textbf{0.19}  & \textbf{0.62}  & \textbf{1.22}  & \textbf{1.92}  & \textbf{2.56}  & \textbf{0.24}  & \textbf{0.72}  & \textbf{1.43}  & \textbf{2.16}  & \textbf{3.11} \\

    \midrule
          &       &       &       &       &       &       &       &       &       &  \\ 
    \midrule
    Dataset & \multicolumn{5}{c}{\textit{left}}     & \multicolumn{5}{c}{\textit{right}} \\
    \cmidrule(r){1-6} \cmidrule(r){7-11} 
    \multirow{2}[1]{*}{Model} & \multicolumn{5}{c}{Horizon (s)}       & \multicolumn{5}{c}{Horizon (s)} \\
    \cmidrule(r){2-6} \cmidrule(r){7-11}         & 1     & 2     & 3     & 4     & 5     & 1     & 2     & 3     & 4     & 5 \\
    S-LSTM \citep{alahi2016social}& 0.77  & 1.68  & 3.04  & 4.67  & 6.59  & 0.69  & 1.97  & 3.81  & 6.17  & 9.09 \\
    S-GAN \citep{gupta2018social}& 0.66  & 1.68  & 3.11  & 4.85  & 6.87  & 0.72  & 1.97  & 3.91  & 6.32  & 9.23 \\
    CS-LSTM \citep{deo2018convolutional}& 0.54  & 1.63  & 3.01  & 4.71  & 6.63  & 0.61  & 2.01  & 3.97  & 6.48  & 9.48 \\
    STDAN \citep{chen2022intention}& \underline{0.35}  & 1.33  & 2.84  & 4.51  & 5.97  & \underline{0.38}  & \underline{1.49}  & 3.46  & 5.87  & 7.93 \\
    WSiP \citep{wang2023wsip}& 0.41  & 1.46  & 2.82  & 4.42  & 6.22  & 0.52  & 1.61  & 3.60  & 5.78  & 8.45 \\
    GaVa \citep{liao2024human}  & 0.45  & \underline{1.22}  & \underline{2.57}  & \underline{3.98}  & \underline{5.95}  & 0.51  & 1.51 & \underline{3.29}  & \underline{5.72}  & {7.82} \\
    \cmidrule(r){1-1} \cmidrule(r){2-6}  \cmidrule(r){7-11} 
    \textbf{HiT (25\%)} & 0.43  & {1.27}  & {2.69}  & {4.21}  & {6.38}  & 0.62  & 1.73 & {3.36}  & {6.41}  & \underline{7.78} \\
    \textbf{HiT} & \textbf{0.31}  &\textbf{1.17}  & \textbf{2.20}  & \textbf{3.63}  & \textbf{5.50}  & \textbf{0.33}  & \textbf{1.38}  & \textbf{2.94}  & \textbf{5.10}  & \textbf{6.72} \\
    \bottomrule
    \end{tabular}%
    }
    \label{Table_manuver}
\end{table}%

\subsection{Comparative Analysis of Model Performance and Model Efficiency}\label{Model Efficiency}
As shown in Table \ref{weight}, HiT reduced the number of parameters by 39.4\% and 49.5\% compared to WSiP and GaVa, respectively. HiT (S) achieved an even greater reduction, using 66.0\% fewer parameters than WSiP whereas simultaneously enhancing performance by at least 20.3\%. These results underscored the efficiency of our models in balancing computational cost and predictive accuracy.

Furthermore, we evaluated the inference speed of HiT against various baselines under identical test conditions using the NGSIM dataset. HiT outperformed all other models in inference speed, running 28.0\% faster than WSiP, the second-fastest model, whereas maintaining the highest prediction accuracy. Specifically, our model surpassed the previous SOTA models GaVa and STDAN, achieving inference speeds that were 23.1\% and 15.4\% faster, respectively. Additionally, Fig. \ref{loss function} illustrates the loss function curves of our model compared to other baselines, revealing that HiT demonstrated consistently faster convergence across all datasets, significantly improving training efficiency over existing approaches.

\begin{table}[htbp]
  \centering
  \caption{Comparison of prediction accuracy and model complexity across different models. The table presents the Average RMSE (m) for various datasets (NGSIM, HighD, RounD, and MoCAD++) and the number of parameters for each model. The inference time was measured for 10 batches of the NGSIM dataset, each with a size of 128, using an Nvidia A40 48G GPU. \textbf{Bold} and \underline{underlined} values are the best and second-best performance.}
  \setlength{\tabcolsep}{2mm}
   \resizebox{0.75\linewidth}{!}{
    \begin{tabular}{ccccccc}
    \toprule
    \multirow{2}[4]{*}{Model} & \multicolumn{4}{c}{Average RMSE (m)} & \multirow{2}[4]{*}{\#Param. (K)} &\multirow{2}[4]{*}{Inference time (s)}\\
\cmidrule{2-5}          & NGSIM & HighD & RounD & MoCAD++ &  \\
    \midrule
    CS-LSTM \citep{deo2018convolutional}& 2.29  & 1.48  & 1.98  & 2.93  &  194.92 &0.22 \\
    CF-LSTM \citep{xie2021congestion}& 1.99  & 1.17  & 1.57  & 1.97  & 387.10 &0.27 \\
    STDAN \citep{chen2022intention}& 1.87  & 0.70  & 1.46  & 1.83  & 486.22 &0.22  \\
    WSiP  \citep{wang2023wsip}& 2.25  & 1.44  & 1.61  & 1.92  & 300.76  &0.25 \\
    GaVa \citep{liao2024human}& \underline{1.65} & \underline{0.39} & - & \underline{1.63} & 360.75 &0.26 \\
    \midrule
    \textbf{HiT (S)} & {1.77}  & {0.41}  & \underline{1.32}  & {1.70} & \textbf{102.40}  & \textbf{0.12} \\
    \textbf{HiT}   & \textbf{1.64} & \textbf{0.25} & \textbf{1.13} & \textbf{1.53} & \underline{182.21} &\underline{0.18}  \\
    \bottomrule
    \end{tabular}%
    }
  \label{weight}%
\end{table}%

\begin{figure}[t]
  \centering
\includegraphics[width=0.95\linewidth]{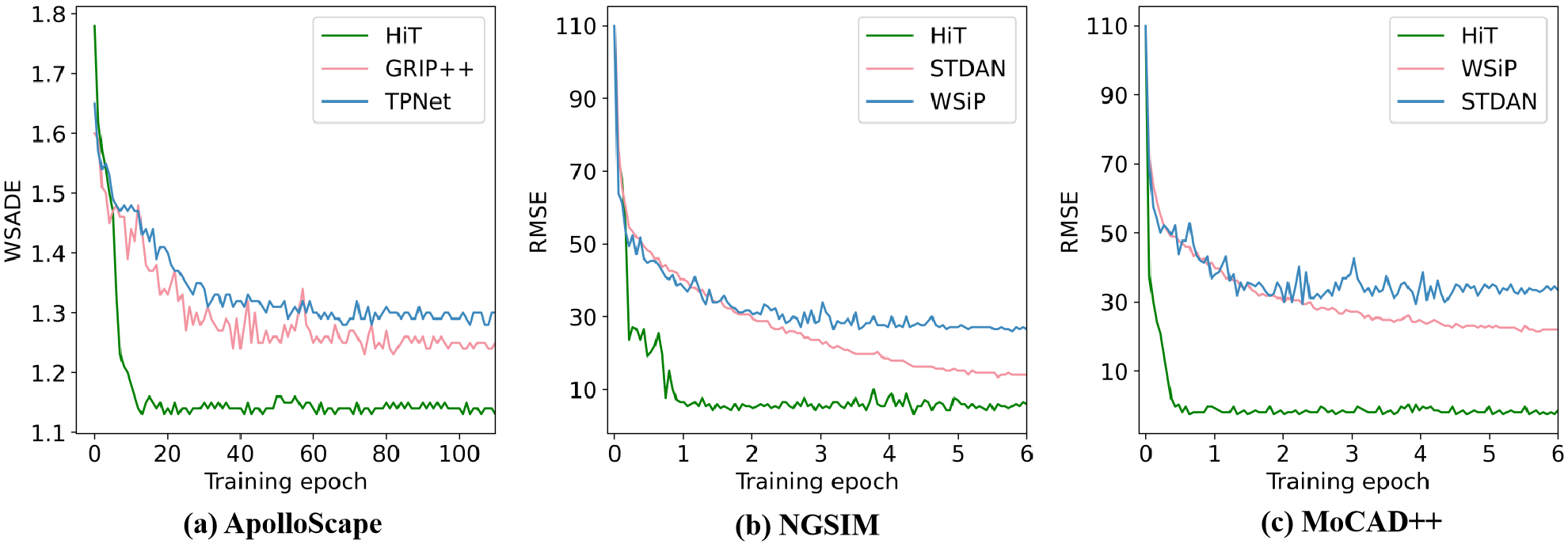} 
  \caption{Visualizations of the loss functions for HiT and selected baselines across four datasets: (a) ApolloScape, (b) NGSIM, and (c) MoCAD++. All models are trained to converge using their optimal hyperparameters. For clarity, we only display the curves up to the point where HiT converges.}
  \label{loss function} 
\end{figure}

The superior convergence rate of HiT can be attributed to several key architectural and training optimizations. Structurally, HiT employed a streamlined design with reduced model complexity, minimizing both trainable parameters and computational overhead, thereby accelerating training. From a training perspective, HiT incorporated the Cosine Annealing Warm Restarts scheduler, which dynamically adjusted the learning rate to avoid local minima and facilitate more efficient exploration and exploitation. Furthermore, the implementation of batch normalization and group normalization stabilized training, mitigating issues such as vanishing and exploding gradients. These enhancements enabled the model to operate with a higher learning rate, further expediting convergence. Additionally, HiT adopted a multi-task training strategy, optimizing both maneuver and trajectory losses simultaneously. This approach allowed the model to learn diverse driving behaviors and multimodal trajectories in parallel, significantly improving training efficiency. Overall, these findings highlighted the effectiveness of our proposed models in achieving a strong balance between inference speed and prediction accuracy, demonstrating their potential for real-time deployment in autonomous driving applications whereas significantly reducing computational costs.

\subsection{Comparative Analysis of Real-time Performance}\label{Real-time Performance}
To evaluate the real-time performance of the HiT model, we compared HiT's inference time against several SOTA models, including STDAN, WSIP, and GAVA, in predicting the future trajectories of 24 traffic agents within complex, multi-agent scenarios.  As summarized in Table \ref{table:performance_inference}, HiT achieved an average inference time of 22 ms per sample, significantly outperforming existing baselines. This performance met the stringent computational requirements of Level 3 autonomous driving systems, which typically operate within a processing range of 20-30 TOPS. Moreover, HiT's computational efficiency ensured seamless deployment on various onboard Neural Processing Units (NPUs) and Data Processing Units (DPUs), such as the Tesla FSD (72 TOPS) and NVIDIA Orin (254 TOPS). Overall, these benchmarks illustrated that HiT achieved an effective balance between high predictive accuracy and computational efficiency, rendering it well-suited for applications in real-world autonomous driving systems.

\section{Ablation Study}
\label{Ablation Study}

\subsection{Ablation Study for Each Module}
Table \ref{Tablem} presents an ablation study evaluating the contributions of key components in our proposed HiT model. We evaluated five variants: Model A, which used the Cartesian coordinate system instead of the polar coordinate system; Model B, which did not include the behavior-aware module; Model C, which used the classical transformer framework in the interaction-aware module instead of the spatial transformer framework; Model D, which excluded the interaction-aware module; Model E, which uses the decoder without the multimodal probabilistic maneuver prediction; and Model F, the complete HiT model.

\begin{table}[tbp]
\centering
\caption{Inference time and accuracy comparison of HiT against SOTA baselines, evaluated on 24 agents using an NVIDIA A40 GPU. \textbf{Bold} values denote the best performance, whereas AVG is the average RMSE metric.}
\setlength{\tabcolsep}{3mm}
\resizebox{0.6\linewidth}{!}
{
\begin{tabular}{ccc}
\toprule
\text{Model} &AVG (m) & Inference Time (ms)   \\
\midrule
STDAN \citep{chen2022intention} &0.70 & 31  \\
WSiP \citep{wang2023wsip} &1.44 &     37   \\
GaVa \citep{liao2024human} &0.39 &   34   \\
\midrule
\textbf{HiT} &\textbf{0.25} & \textbf{22}  \\
\bottomrule
\end{tabular}}
\label{table:performance_inference}
\end{table}

\begin{table}[htbp]
  \centering
  \caption{Different methods and components of ablation study.}\label{Tablem}
   \setlength{\tabcolsep}{3mm}
  \resizebox{0.6\linewidth}{!}{
    \begin{tabular}{ccccccc}
    \toprule
    \multirow{2}[4]{*}{Components} & \multicolumn{6}{c}{Ablation methods} \\
\cmidrule{2-7}          & A     & B     & C     & D     & E  &F\\
    \midrule
    Polar coordinate system & \ding{56} & \ding{52} & \ding{52} & \ding{52} & \ding{52} & \ding{52} \\
    Behavior-aware module  & \ding{52} & \ding{56} &  \ding{52} &\ding{52} & \ding{52}  & \ding{52} \\
    Spatial transformer framework & \ding{52} & \ding{52} & \ding{56}& \ding{56}  & \ding{52} & \ding{52} \\
    Interaction-aware module & \ding{52} & \ding{52} & \ding{52} & \ding{56} & \ding{52}  & \ding{52} \\
     Multimodal Decoder & \ding{52} & \ding{52} & \ding{52} & \ding{52}& \ding{56}  & \ding{52} \\
    \bottomrule
    \end{tabular}}%
\end{table}%

We evaluated these models' performance using RMSE on the NGSIM, HighD, RounD, and MoCAD++ datasets, with the results reported in Table \ref{Table4}. Compared to model F, which includes all five components, the performance of all ablation methods declined significantly. Model F consistently excelled in the evaluation metrics, underscoring the collective value of the components in achieving optimal performance. Overall, method B and method D performed the worst, highlighting the importance of the Behavior-aware module and the Interaction-aware module. This quantitatively demonstrates the significant impact of behavior-awareness and interaction-awareness—two critical factors influencing driver decision-making—on trajectory prediction. Furthermore, omitting the polar coordinate system in method A led to a noticeable decline in predictive performance. The individual contributions of the proposed spatial transformer framework and multimodal decoder are also evident from the performance drop in method C and method E, respectively.

Specifically, the inclusion of the Behavior-aware module significantly improved performance by capturing driver behaviors, which is crucial for accurate trajectory prediction. By considering the behavior of surrounding vehicles, HiT can more accurately predict the ego vehicle's trajectory, reflecting the importance of behavior in the human decision-making process \citep{baron2000thinking}. When analyzing the ablation results on the RounD and MoCAD++ datasets, we observed that the decline in performance for method A—where the Cartesian coordinate system replaced the polar coordinate system—was more pronounced than on the HighD and NGSIM highway datasets. This indicates that the proposed polar coordinate-based pooling mechanism is particularly important for enhancing predictive capabilities in unstructured environments.

In summary, these results demonstrate the significance of each of the five components in achieving optimal model performance and emphasize the value of considering both agent behavior and relative spatial interaction relationships in trajectory prediction for autonomous vehicles.

\begin{table}[t]
  \centering
  \caption{Ablation results for models on NGSIM, HighD, RounD, and MoCAD++ datasets. Bold values denote the best performance in each category. (Evaluation metric: RMSE (m))}
\setlength{\tabcolsep}{5mm}
\resizebox{0.7\linewidth}{!}{
    \begin{tabular}{c|ccccccc}
    \toprule
    \multicolumn{1}{c}{\multirow{2}[4]{*}{Dataset}} & \multirow{2}[4]{*}{Time (s)} & \multicolumn{6}{c}{ Model } \\
\cmidrule{3-8}    \multicolumn{1}{c}{} &      & A     & B     & C     & D     & E  &F \\ 
  \hline
    \multirow{5}[2]{*}{NGSIM} 
          & 1     &0.55  & 0.73  & 0.45  & 0.63 &0.50 & \textbf{0.38} \\
          & 2     & 1.24  & 1.89  &1.16  & 1.57  &1.19& \textbf{0.90} \\
          & 3     & 1.80  & 2.14  & 1.58  & 2.01 &1.72 & \textbf{1.42} \\
          & 4     & 2.38  & 2.86 & 2.27  & 2.65  &2.34& \textbf{2.08} \\
          & 5     & 3.25  & 3.87  & 3.02  & 3.56  &3.11& \textbf{2.87} \\
 \hline
     \multirow{5}[2]{*}{HighD} 
          & 1     &0.17  & 0.36  & 0.14  & 0.26  &0.18 &\textbf{0.08} \\
          & 2     & 0.26  & 0.50  & 0.20 &  0.47 &0.27 & \textbf{0.13} \\
          & 3     & 0.32  & 0.63  & 0.28 &  0.53 &0.36 & \textbf{0.22} \\
          & 4     & 0.51  & 0.95 & 0.58 & 0.81 & 0.61& \textbf{0.39} \\
          & 5     & 0.79  & 1.27  & 0.84 &  1.19 & 0.89 &\textbf{0.61} \\
 \hline
    \multirow{5}[0]{*}{RounD} 
          & 1     & 0.62  & 0.81  & 0.39  & 0.69 & 0.31 & \textbf{0.21} \\
          & 2     & 1.38  & 1.51  & 0.69  & 1.38 & 0.64 & \textbf{0.52} \\
          & 3     & 2.24  & 2.39  & 1.70  & 2.32 & 1.61& \textbf{1.40} \\
          & 4     & 3.17  & 3.31  & 2.68  & 3.21 & 2.65 & \textbf{2.41} \\
            \hline
      \multirow{5}[2]{*}{MoCAD++}     
          & 1     & 0.62  & 0.79  & 0.47  & 0.68 & 0.42& \textbf{0.32} \\
          & 2     & 1.12  & 1.27  & 0.92  & 1.21 &0.85 & \textbf{0.75}  \\
          & 3     & 1.82  & 2.01  & 1.61  & 1.93 &1.52 & \textbf{1.36} \\
          & 4     & 2.72  & 3.04 & 2.31  & 2.89 &2.38 & \textbf{2.18} \\
          & 5     & 3.36  & 3.68  & 3.15  & 3.52 &3.06 & \textbf{2.82} \\
    \bottomrule
    \end{tabular}%
    }
 \label{Table4}
\end{table}%

\begin{table}[htb]
  \centering
  \caption{Different methods and components of the ablation study.}\label{Tablem_bahavior}
   \setlength{\tabcolsep}{3mm}
  \resizebox{0.6\linewidth}{!}{
    \begin{tabular}{ccccccc}
    \toprule
    \multirow{2}[4]{*}{Components} & \multicolumn{6}{c}{Ablation methods} \\
\cmidrule{2-7}          & G     & H     & I     & J     & K &L\\
    \midrule
    Dynamic degree centrality & \ding{56} & \ding{52} & \ding{56} & \ding{52} & \ding{52} & \ding{52} \\
    Dynamic closeness centrality  & \ding{52} & \ding{56} &  \ding{56} &\ding{52} & \ding{52}  & \ding{52} \\
    q-ROFWEBM operator & \ding{52} & \ding{52} & \ding{52} & \ding{56} & \ding{52}  & \ding{52} \\
     Hypergraph behavior learning & \ding{52} & \ding{52} & \ding{52} & \ding{52}& \ding{56}  & \ding{52} \\
    \bottomrule
    \end{tabular}}%
\end{table}%

\subsection{Ablation Study for Behavior-aware Module}
As depicted in Table \ref{Tablem_bahavior}, we conducted experiments using five different models: Model G, which used the classical degree centrality, instead of the dynamic degree centrality; Model H, which used the classical closeness centrality, instead of the dynamic closeness centrality; Model I, which used the classical closeness centrality and classical degree centrality; Model J, which used the classical degree centrality, instead of the dynamic degree centrality; Model K, which used the graph behavior learning, instead of the hypergraph approach; and Model L, the HiT model with complete Behavior-aware module.

According to the data presented in Table \ref{Table4_behavior}, the removal of components within the behavior-aware module consistently impacts prediction accuracy. Among individual components, the dynamic degree centrality has the most significant effect on prediction accuracy across all datasets, indirectly validating its strong capability in modeling driver behavior. Replacing the hypergraph with a graph neural network for behavior learning also led to a decrease in accuracy, indicating that the hypergraph effectively captures the higher-order correlations associated with driver behavior. The elimination of other components similarly resulted in a drop in prediction accuracy, underscoring the value and necessity of each component. Moreover, we observed that method I, which simultaneously removes both dynamic degree centrality and closeness centrality, exhibits a more pronounced decline in prediction accuracy. This decline suggests that the combined use of these two centrality measures provides a significantly stronger modeling capability for driver behavior than either measure alone, emphasizing the importance of a multi-metric design.

\begin{table}[t]
  \centering
  \caption{Ablation results for different models on NGSIM, HighD, RounD, and MoCAD++ datasets. Notably, the RMSE (m) is used as the evaluation metric. Bold values denote the best performance in each category.}
\setlength{\tabcolsep}{4mm}
\resizebox{0.75\linewidth}{!}{
    \begin{tabular}{c|ccccccc}
    \toprule
    \multicolumn{1}{c}{\multirow{2}[4]{*}{Dataset}} & \multirow{2}[4]{*}{Time (s)} & \multicolumn{6}{c}{ Model } \\
\cmidrule{3-8}    \multicolumn{1}{c}{} &       & G     & H     & I     & J     & K &L \\
   \hline
    \multirow{5}[2]{*}{NGSIM} 
          & 1     & 0.43  & 0.40  & 0.44 & 0.42 & 0.41 & \textbf{0.38} \\
          & 2     & 1.02  & 0.92  & 1.08 & 0.99 & 0.94 & \textbf{0.90} \\
          & 3     & 1.51  & 1.44  & 1.52 & 1.49 & 1.47 & \textbf{1.42} \\
          & 4     & 2.17  & 2.10  & 2.19 & 2.15 & 2.12 & \textbf{2.08} \\
          & 5     & 2.94  & 2.89  & 2.97 & 2.93 & 2.91 & \textbf{2.87} \\
    \hline
     \multirow{5}[2]{*}{HighD} 
          & 1     &0.14  & 0.10  & 0.15  & 0.13  &0.11 &\textbf{0.08} \\
          & 2     & 0.20  & 0.15  & 0.23 &  0.18 &0.16 & \textbf{0.13} \\
          & 3     & 0.28  & 0.24  & 0.29 &  0.27 &0.26 & \textbf{0.22} \\
          & 4     & 0.46  & 0.41 & 0.47 & 0.44 & 0.42& \textbf{0.39} \\
          & 5     & 0.69  & 0.64  & 0.72 &  0.67 & 0.65 &\textbf{0.61} \\
 \hline
    \multirow{5}[0]{*}{RounD} 
        & 1     & 0.28  & 0.23 & 0.29  & 0.26 &0.25 & \textbf{0.21} \\
          & 2     & 0.59  & 0.54  & 0.61  & 0.57 &0.55 & \textbf{0.52} \\
          & 3     & 1.51  & 1.42  & 1.52  & 1.49  &1.46& \textbf{1.40} \\
          & 4     & 2.51  & 2.44  & 2.54  & 2.49 &2.46 & \textbf{2.41} \\
  \hline
      \multirow{5}[2]{*}{MoCAD++}     
          & 1     &0.39  & 0.34  & 0.40  & 0.37 & 0.36& \textbf{0.32} \\
          & 2     & 0.81  & 0.77  & 0.83  & 0.80 &0.78 & \textbf{0.75}  \\
          & 3     & 1.45  & 1.38  & 1.46  & 1.43 &1.41 & \textbf{1.36} \\
          & 4     & 2.27  & 2.20 & 2.29  & 2.25 &2.22 & \textbf{2.18} \\
          & 5     & 2.92  & 2.86  & 2.96  & 2.89 &2.87 & \textbf{2.82} \\
    \bottomrule
    \end{tabular}%
    }
 \label{Table4_behavior}
\end{table}%

\end{APPENDIX}

\end{document}